\documentclass{article}
\usepackage{natbib}
\setcitestyle{numbers,square}
\usepackage[final]{neurips_2022}
\usepackage[utf8]{inputenc} % allow utf-8 input
\usepackage[T1]{fontenc}    % use 8-bit T1 fonts
\usepackage{url}            % simple URL typesetting
\usepackage{booktabs}       % professional-quality tables
\usepackage{amsfonts}       % blackboard math symbols
\usepackage{nicefrac}       % compact symbols for 1/2, etc.
\usepackage{microtype}      % microtypography
\usepackage{algorithm}%
\usepackage{algpseudocode}
\usepackage{amsfonts,amsmath,amssymb,amsthm}
\usepackage{dsfont}
\usepackage{lipsum}
\usepackage{wrapfig}
\usepackage{subfigure}
\usepackage{graphicx}
\usepackage{vcell}
\usepackage{makecell}
\usepackage{multirow}
\usepackage{multicol}
\usepackage[table,xcdraw,dvipsnames]{xcolor}
\usepackage{enumitem}
\usepackage{tikz}
\usepackage[font=small]{caption}
\usepackage{xcolor}
\usepackage{tcolorbox}
\usepackage{bbm} % indicator function
\usepackage{hyperref}
\usepackage{csquotes}
\usepackage{cleveref}
\hypersetup{
 colorlinks=True,
 linkcolor=blue,
 citecolor=blue,
 urlcolor=blue}

\definecolor{mine}{RGB}{205, 232, 248}%

\newtheorem{assumption}{Assumption}
\newtheorem{theorem}{Theorem}

\title{A Policy-Guided Imitation Approach for \\ Offline Reinforcement Learning}
\author{
Haoran Xu$^\spadesuit$\footnotemark[1]  \quad
Li Jiang$^\clubsuit$\footnotemark[1]  \quad
Jianxiong Li$^\clubsuit$  \quad
Xianyuan Zhan$^{\clubsuit,\diamondsuit}$\\
$^\spadesuit$JD Technology, Beijing, China\\
$^\clubsuit$Tsinghua University, Beijing, China\\ 
$^\diamondsuit$Shanghai AI Laboratory, Shanghai, China\\
\texttt{\{ryanxhr,jiangli3859\}@gmail.com}
}

\begin{document}

\maketitle

\renewcommand{\thefootnote}{\fnsymbol{footnote}}
\footnotetext[1]{Equal contribution. Correspondence to Haoran Xu, Xianyuan Zhan.}

\begin{abstract}
Offline reinforcement learning (RL) methods can generally be categorized into two types: RL-based and imitation-based. RL-based methods could in principle enjoy out-of-distribution generalization but suffer from erroneous off-policy evaluation. Imitation-based methods avoid off-policy evaluation but are too conservative to surpass the dataset. In this study, we propose an alternative approach, inheriting the training stability of imitation-style methods while still allowing logical out-of-distribution generalization. We decompose the conventional reward-maximizing policy in offline RL into a guide-policy and an execute-policy. During training, the guide-policy and execute-policy are learned using only data from the dataset, in a supervised and decoupled manner. During evaluation, the guide-policy guides the execute-policy by telling where it should go so that the reward can be maximized.
% , serving as the \textit{Prophet}. 
By doing so, our algorithm allows \textit{state-compositionality} from the dataset, rather than \textit{action-compositionality} conducted in prior imitation-style methods. We dumb this new approach Policy-guided Offline RL (\texttt{POR}). \texttt{POR} demonstrates the state-of-the-art performance on D4RL, a standard benchmark for offline RL. We also highlight the benefits of \texttt{POR} in terms of improving with supplementary suboptimal data and easily adapting to new tasks by only changing the guide-policy. 
Code is available at \url{https://github.com/ryanxhr/POR}.
\end{abstract}

\section{Introduction}
% Offline RL: what and why.
Offline RL, also known as batch RL, allows learning policies from previously collected data, without online interactions \cite{lange2012batch,levine2020offline}. It is a promising area for bringing RL into real-world domains, such as robotics \cite{kalashnikov2021mt}, healthcare \cite{tang2021model} and industrial control \cite{zhan2022deepthermal}. In such scenarios, arbitrary exploration with untrained policies is costly or dangerous, but sufficient prior data is available. 
While most off-policy RL algorithms are applicable in the offline setting by filling the replay buffer with offline data, improving the policy beyond the level of the behavior policy often entails querying the value function (i.e., Q function) about values of actions that were not seen in the dataset. Those out-of-distribution actions can be deemed as adversarial examples of the Q function \cite{kumar2020conservative}, which cause extrapolation error of the Q-function. The error will be accumulated by the deadly triad issue \cite{van2018deep}, propagate across the state-action space through the iterative dynamic programming procedure \cite{brandfonbrener2021offline}, and can not be eliminated without requiring a growing batch of online samples \cite{levine2020offline}.
% Therefore, the main benefit of offline RL, the lack of environment interaction, is also what makes it a challenging task. 

% Offline RL: how (prior work).
To alleviate this issue, prior model-free offline RL methods typically add a behavior regularization term to the policy improvement step, to limit how far it deviates from the behavior policy. This can be achieved explicitly by calculating some divergence metrics \cite{kumar2019stabilizing,wu2019behavior,nair2020accelerating,fujimoto2021minimalist}, or implicitly by regularizing the learned value functions to assign low values to out-of-distribution actions \cite{kumar2020conservative,kostrikov2021offline,an2021uncertainty,bai2021pessimistic}.
Nevertheless, this imposes a trade-off between accurate value estimation (more behavior regularization) and maximum policy improvement (less behavior regularization).
% Nevertheless, this imposes a trade-off between how much the policy improves and how vulnerable it is to misestimation due to distributional shift.
To avoid this problem, recently a branch of methods bypass querying the values of unseen actions by performing some kind of imitation learning on the dataset\footnote{The core difference between RL-based and imitation-based methods is that RL-based methods learn a value function of policy $\pi$ while imitation-based methods don't.}.
This can be achieved by filtering trajectories based on their return \cite{peng2019advantage,chen2020bail}, or reweighting transitions based on how advantageous they could be under the behavior policy \cite{brandfonbrener2021offline,kostrikov2021iql}, or just be conditioned on some variables without any dataset reweighting \cite{chen2021decision,emmons2021rvs}. 
Although imitation-style methods enjoy a stable training process and are able to effectively perform multi-step dynamic programming by assigning the proper reweighting weight or conditioned variable \cite{kostrikov2021iql,emmons2021rvs}, they only allow \textit{action-compositionality} from the dataset, lose the ability to surpass the dataset by out-of-distribution generalization, which only appears in RL-based methods.

% Offline RL: how and why (our paper).
In this work, we propose an alternative approach. We aim at inheriting the training stability of imitation-style methods while still allowing logical out-of-distribution generalization.
% Apart from the typical reward-maximizing policy (execute-policy), we introduce an additional guide-policy. During training, the guide-poicy and execute-policy are learned using only data from the dataset, in a supervised and decoupled manner. During evaluation, the guide-policy guides the execute-policy by telling where it should go so that the reward can be maximized, serving as \textit{The Prophet}. By doing so, our algorithm allows \textit{state-compositionality} on the dataset, rather than \textit{action-compositionality} conducted in prior work.
To do so, we propose Policy-guided Offline RL (\texttt{POR}), an algorithm that employs two policies to solve tasks: a guide-policy and an execute-policy. The job of the guide-policy is to learn the optimal next state given the current state, and the job of the execute-policy is to learn how different actions can produce different next states, given the current state.
By this manner, we decompose the original task of RL (i.e., reward-maximizing) into two distinct yet complementary tasks, and this decomposition makes each task much easier to be solved.
During evaluation, the guide-policy serves as a guide for the execute-policy by telling it where to go so that the reward can be maximized. 
By doing so, our algorithm allows \textit{state-compositionality} from the dataset, rather than \textit{action-compositionality} conducted in prior work, which owns logical out-of-distribution generalization.

% We term this new approach Policy-guided Offline RL (\texttt{POR}).
Our method is easy to implement by only adding the learning of a guide-policy, the training stage of the guide-policy and execute-policy are decoupled, using in-sample learning from the dataset.
We test \texttt{POR} in widely-used D4RL offline RL benchmarks and demonstrates the state-of-the-art performance, especially on low-quality datasets that require out-of-distribution generalization to achieve a high score. 
We also show that by decoupling the learning of guide-policy and execute-policy, we can enhance the guide-policy with supplementary suboptimal data, or re-learn the guide-policy to adapt to new tasks with different reward functions, without changing the execute-policy.

\section{Related Work}
\noindent \textbf{RL-based offline approach} \quad
A large portion of offline RL methods is RL-based. RL-based methods typically augment existing off-policy methods (e.g., Q-learning or actor-critic) with a behavior regularization term. The primary ingredient of this class of methods is to propose various policy regularizers to ensure that the learned policy does not stray too far from the behavior policy. 
These regularizers can appear explicitly as divergence penalties~\cite{wu2019behavior,kumar2019stabilizing,fujimoto2021minimalist}, implicitly through weighted behavior cloning~\cite{wang2020critic,peng2019advantage,nair2020accelerating}, or more directly through careful parameterization of the policy~\cite{fujimoto2018addressing,zhou2020latent}. 
Another way to apply behavior regularizers is via modification of the critic learning objective to incorporate some form of regularization to encourage staying near the behavioral distribution and being pessimistic about unknown state-action pairs~\cite{nachum2019algaedice,kumar2020conservative,kostrikov2021offline,xu2022constraints}. 
There are also several works incorporating behavior regularization through the use of uncertainty. The uncertainty quantification can be done via Monte Carlo dropout \cite{wu2021uncertainty} or explicit ensemble models \cite{bai2021pessimistic,yu2020mopo,kidambi2020morel,zhan2022model}.
Note that the behavior regularization weight is crucial in RL-based methods. With a small weight, RL-based methods could in principle enjoy out-of-distribution generalization since they perform true dynamic programming. However, a small weight will make erroneous off-policy evaluation due to distribution shift and the policy is extremely vulnerable to the value function misestimation.
% Nevertheless, this imposes a trade-off between accurate value estimation (more behavior regularization) and maximum policy improvement (less behavior regularization)

\noindent \textbf{Imitation-based offline approach} \quad
Another line of methods, on the contrary, performs some kind of imitation learning on the dataset, without the need to do off-policy evaluation.
When the dataset is good enough or contains high-performing trajectories, we can simply clone the actions observed in the dataset \cite{pomerleau1989alvinn}, or perform some kind of filtering or conditioning to extract useful transitions. 
For instance, recent work filters trajectories based on their return \cite{chen2020bail,peng2019advantage}, or directly filters individual transitions based on how advantageous these could be under the behavior policy and then clones them \cite{brandfonbrener2021offline,kostrikov2021iql,brandfonbrener2021quantile,gulcehre2021regularized,xu2021offline}. 
Conditioned BC methods are based on the idea that every transition or trajectory is optimal when conditioned on the right variable \cite{ghosh2020learning,emmons2021rvs}. In this way, after conditioning, the data becomes optimal given the value of the conditioned variable. 
The conditioned variable could be the cumulative return \cite{chen2021decision,janner2021offline,kumar2019reward}, or the goal information if provided \cite{lynch2020learning,ghosh2020learning,emmons2021rvs,yang2022rethinking}.
A recent study \cite{emmons2021rvs} shows that goal-conditioning can be super useful in D4RL AntMaze datasets. However, this introduces the assumption that we have some prior information about the structure of the task, which is beyond the standard offline RL setup. 
% We think an exciting direction for future work would be to remove this assumption by automating the learning of the goal space.

Our method can be viewed as a combination of RL-based and imitation-based methods. We adopt an imitation-style manner to train the goal-policy and execute-policy. During evaluation, however, following the guide-policy, the execute-policy could produce out-of-distribution actions by leveraging the generalization capacity of the function approximator.
% The power of supervised learning. 
% This to estimate the value of the best available action at a given state without ever directly querying a Q-function with this unseen action.

% \noindent \textbf{Goal-conditioned RL} \quad
% In our learning objective, xxxx. This draws the connection to the Goal-conditioned RL literature.
% Subgoal generation.
% We refer the readers to \cite{liu2022goal} for a comprehensive review.

\section{Preliminaries}

\noindent \textbf{Offline RL} \quad
We consider the standard fully observed Markov Decision Process (MDP) setting~\cite{sutton1998introduction}. 
An MDP can be represented as $\mathcal{M}=(\mathcal{S}, \mathcal{A}, T, r, \rho, \gamma)$ where $\mathcal{S}$ is the state space, $\mathcal{A}$ is the action space, $T(\cdot | s, a)$ is the transition probability distribution function, $r(s, a)$ is the reward function, $\rho$ is the initial state distribution and $\gamma$ is the discount factor, we assume $\gamma \in (0, 1)$ in this work. 
% The goal of RL is to find a policy $\pi(\cdot | s)$ that maximizes the expected cumulative discounted reward starting from $\rho$ as 
The goal of RL is to find a policy $\pi(a|s): \mathcal{S} \times \mathcal{A} \rightarrow[0,1]$ that maximizes the expected cumulative discounted reward (or called return) along a trajectory as
\begin{align}
\label{eq_rl}
\max_{\pi} \ \mathbb{E} \left[\sum_{t=0}^{\infty} \gamma^{t} r\left(s_{t}, a_{t}\right) \bigg| s_{0}=s, a_{0}=a, s_{t} \sim T\left(\cdot | s_{t-1}, a_{t-1}\right), a_{t} \sim \pi\left(\cdot|s_{t}\right) \text { for } t \geq 1\right].
\end{align}
In this work, we focus on the offline setting. The goal is to learn a policy from a fixed dataset $\mathcal{D}=\left\{\tau^{i}=\left(\left(s_{0}^{i}, a_{0}^{i}, s_{0}^{\prime i}, r_{0}^{i}\right),\left(s_{1}^{i}, a_{1}^{i}, s_{1}^{\prime i}, r_{1}^{i}\right), \cdots,\left(s_{H}^{i}, a_{H}^{i}, s_{H}^{\prime i}, r_{H}^{i}\right)\right)\right\}_{i=1}^{N}$ consisting of trajectories that are collected by different policies, where $H$ is the time horizon.
The dataset can be heterogenous and suboptimal, we denote the underlying behavior policy of $\mathcal{D}$ as $\mu$, which represents the conditional distribution $p(a|s)$ observed in the dataset.

\noindent \textbf{RL via Supervised Learning} \quad
Conventional RL methods generally either compute the derivative of (\ref{eq_rl}) with respect to the policy directly via policy gradient methods \cite{schulman2015trust,kakade2001natural}, or estimate a value function or Q-function by means of TD learning \cite{watkins1992q}, or both \cite{silver2014deterministic, haarnoja2018soft}.
In the \textit{RL via Supervised Learning} framework, we avoid the complex and potentially high-variance policy gradient estimators, as well as the complexity of temporal difference learning. Instead, we perform conditioned behavior cloning on some extra information, such as goal state, cumulative return, or language description \cite{lynch2020learning,ghosh2020learning}. 
When applying the framework to the offline RL setting, we can take the offline dataset $\mathcal{D}$ as input and find the outcome-conditioned policy $\pi$ that optimizes
\begin{align}
\max_{\pi} \ \mathbb{E}_{\tau \sim \mathcal{D}, t \sim \operatorname{Unif}(1, H), \omega \sim g\left(\cdot | \tau_{t:H} \right) } \left[\log \pi\left(a_{t} | s_{t}, \omega \right)\right].
% \max _{\pi} \sum_{\tau \in \mathcal{D}} \sum_{1 \leq t \leq|\tau|} \mathbb{E}_{\omega \sim f\left(\omega \mid \tau_{t: H}\right)}\left[\log \pi_{\theta}\left(a_{t} \mid s_{t}, \omega\right)\right],
\end{align}
where $\omega$ is an outcome conditioned on the remaining trajectory, i.e., $\omega \sim g\left(\cdot | \tau_{t: H}\right)$, and we use $\tau_{i: j}=\left(s_{i}, a_{i}, \ldots, s_{j}\right)$ to denote a fragment of the trajectory.
% It is also possible to condition on other information, such as the parameters of a reward function inferred via inverse RL [10] or the parameters of a task, such as turning left or right [6].

\section{Policy-guided Offline Reinforcement Learning}
We now continue to describe our proposed approach, \texttt{POR}.
To have a clear understanding of the out-of-distribution generalization of \texttt{POR}, we begin with a motivating example which demonstrates that by allowing state-compositionality on the dataset, the agent is able to generalize to a much better policy than doing action-compositionality.
Then we describe how we accomplish this idea in the more general case and introduce the algorithm details of \texttt{POR}. 
Finally, we give a theoretical analysis of \texttt{POR}, we are interested to see how close the policy induced by \texttt{POR} and the optimal policy could be.

\subsection{A Motivating Example}
\begin{wrapfigure}{r}{0.45\textwidth}
% \vspace{-0.5cm}
\centering
\includegraphics[width=0.45\textwidth]{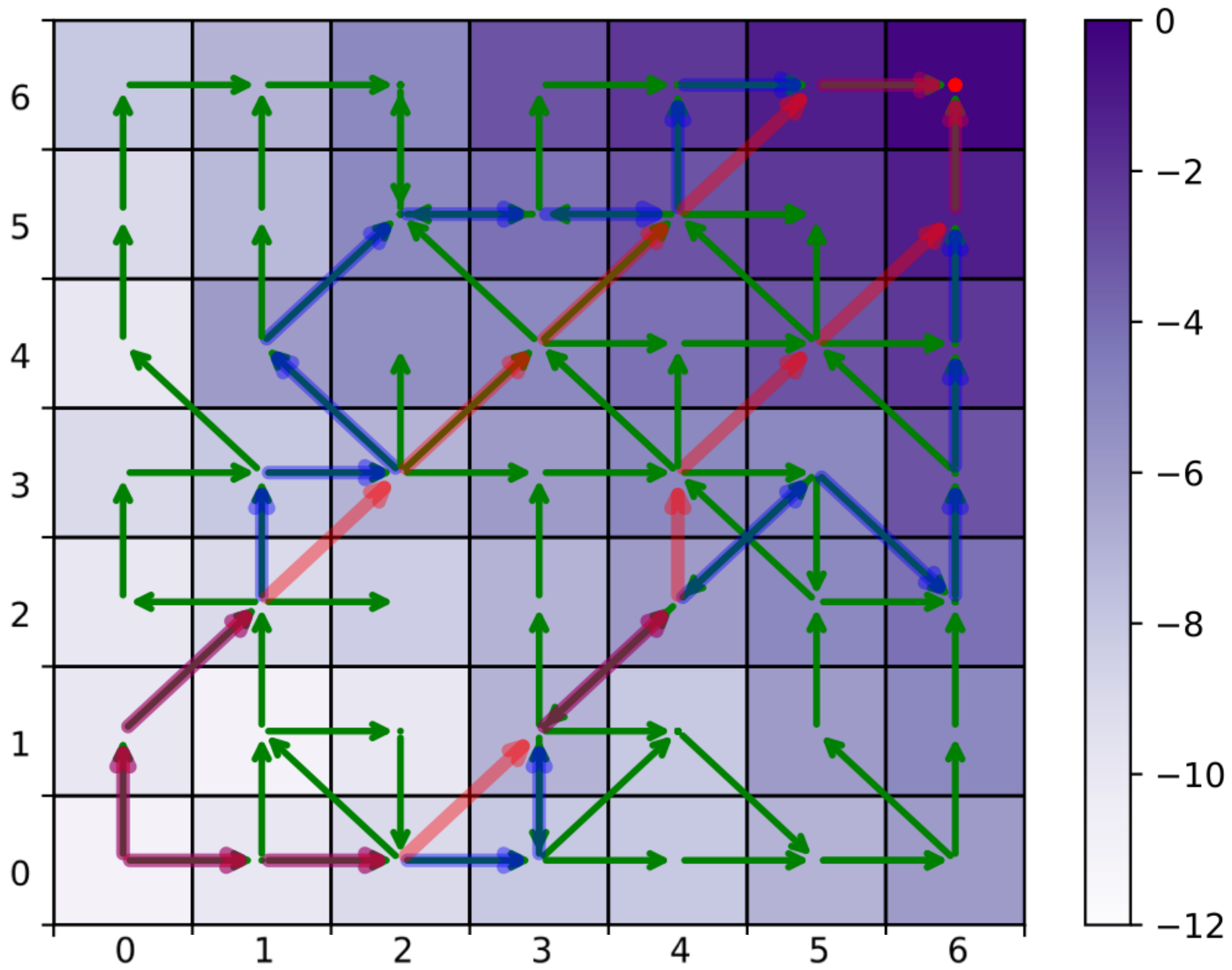}
\caption{A toy example in the grid world that requires navigation from the start location (bottom-left corner) to the goal location (top-right corner). The offline dataset is colored with green, trajectories obtained from action-stitching method and state-stitching method are colored with blue and red, respectively. Note that both action-stitching and state-stitching methods have two equal-valued action choices at the start location ($\uparrow$ and $\rightarrow$), resulting in two different trajectories in both blue and red lines.}
\vspace{-0.4cm}
\label{fig:toy case}
\end{wrapfigure}

We first use a simple, didactic toy example to illustrate the importance of state-stitching for offline RL. 
Let's consider the navigation task shown in Figure \ref{fig:toy case}, where the goal is to find the shortest path from the starting location (bottom-left corner) to the goal location (top-right corner) in the grid world. 
% This is directly representative of several real-world decision-making scenarios in mobile robot navigation.
The state and action space in this task are discrete, the state consists of the $(x,y)$ coordinate and the agent could choose to move to any of its nearby 8 grids (i.e., the action space is 8).
The agent receives a reward of $0$ for entering the goal state and $-1$ for all other transitions.
The offline dataset (green-arrowed lines) consists of two suboptimal trajectories from the start location to the goal location, plus transitions of taking random actions at random states.

We use the grid color (more saturated to purple means higher value) to show the value function $V(s)$ computed from the offline dataset.
We compare two methods that use the value function differently. At a state $s_b \sim \mathcal{D}$, the first method chooses the action $a_b$ in the dataset that leads to $s^{\prime}_{b}$ with the highest $V(s^{\prime}_{b})$; the second method chooses the action $a$ that leads to $s^{\prime}$ in the dataset with the highest $V(s^{\prime})$ , note that $a$ is not necessarily to be within the dataset.
The first method is doing action "stitching" in the dataset while the second method is doing state "stitching". Although both methods are able to "stitch" suboptimal trajectories together from the dataset and successfully travel from the start location to the goal location, the action-stitching method takes more steps (11 steps) compared with the state-stitching method (7 or 8 steps), as shown in Figure \ref{fig:toy case}.

The action-stitching method introduced above is actually an abstract methodology of imitation-based methods, which imitate the transitons in the dataset unequally with different weight choices.
However, these methods are too conservative to find the optimal trajectory, especially when there does not exist one in the offline dataset. To generate better-than-dataset trajectories (e.g., red-arrowed lines), the agent needs to take logical out-of-distribution actions.

Note that in the toy example, we give the agent the freedom to choose \textbf{any} action that leads to the highest-value state in the dataset. This allows the most out-of-distribution generalization and is realizable in some simple tasks (e.g., discrete state and action space) with high action coverage, for instance, in the toy example, given historical $(s,a,s^{\prime})$ results at $(5, 4)$, the agent can successfully recognize that taking upper-right will arrive at $(6, 5)$. However, in the more general case, especially the continuous state and action space setting, this may make the agent do erroneously-generalized actions as the same state can hardly be observed twice.
To perform logical out-of-distribution generalization, we need a \textit{Prophet}, i.e., the guide-policy, to help the agent by telling it which state it should (high reward) and can (logical generalization) go to.
It turns out that the involvement of the guide-policy also brings some theoretical meaning, we will discuss that in Section \ref{sec: theory}.

\subsection{Learning the Guide-Policy}
Recall that the goal of the guide-policy is to guide the execute-policy about which state $s$ it should go to. To accomplish that, we train a state value function $V(s): \mathcal{S} \rightarrow \mathbb{R}$. 
The training of $V$ uses only $(s, s')$ samples in the offline dataset, it  doesn't suffer from overestimation because there're no out-of-distribution actions involved. 
% However, this only learn the value of behavior policy $\mu$ \cite{brandfonbrener2021offline}. 
To approximate the optimal value function in the dataset, we adopt tricks from recent work \cite{kostrikov2021iql,ma2022offline} by giving the $\ell_{2}$ loss with a different weight using expectile regression, yielding the following asymmetric $\ell_{2}$ loss
\begin{equation}
    % \min_{\phi} \ \mathbb{E}_{(s, r, s') \sim \mathcal{D}} \left[L_{2}^{\tau}\left(r + \gamma V_{\phi'}(s') - V_{\phi}(s)\right)\right], \text { where } L_{2}^{\tau}(u)=|\tau-\mathbbm{1}(u<0)| u^{2}.
    \min_{\phi} \ \mathbb{E}_{(s, r, s') \sim \mathcal{D}} \bigg[ \Big|\tau -\mathds{1} \Big(r + \gamma V_{\phi'}(s') - V_{\phi}(s) < 0 \Big) \Big|  \Big(r + \gamma V_{\phi'}(s') - V_{\phi}(s) \Big)^2 \bigg].
    \label{eq: update V}
\end{equation}
It can be seen that when $\tau=1 / 2$, this operator is reduced to Bellman expectation operator, while when $\tau \rightarrow 1$, this operator approaches Bellman optimality operator. 
Note that our learning objective bears similarity with IQL \cite{kostrikov2021iql}, however, here we aim to learn a state value function while IQL aims to learn a state-action value function so as to extract the policy. To do this, IQL needs to learn both $Q$ and $V$ ($V$ is used to isolate the effect of state and approximate the expectile of $Q$ only with respect to the action distribution), the training of $V$ and $Q$ are coupled and may affect each other.

Inspired by recent work \cite{xu2023offline} that enables in-sample learning via implicit value regularization, we additionally propose a sparse value learning objective for $V$, which is a more principled way to approximate the optimal $V$ in the dataset while preventing from distributional shift, by
\begin{equation}
    \min_{\phi} \ \mathbb{E}_{(s, r, s') \sim \mathcal{D}} \bigg[ \mathds{1} \Big( 1 + \frac{r + \gamma V_{\phi'}(s') - V_{\phi}(s)}{2\tau} >0 \Big) \Big(1 + \frac{r + \gamma V_{\phi'}(s') - V_{\phi}(s)}{2\tau} \Big)^2 + \frac{V_{\phi}(s)}{\tau} \bigg].
    \label{eq: update V sql}
\end{equation}
We denote this version of \texttt{POR} as \texttt{POR-sparse}. In \texttt{POR-sparse}, $\tau$ controls the trade-off between approximating the optimal $V$ and reducing the approximation error, a smaller $\tau$ drives towards a better $V$ whereas introducing larger approximation error, and vice versa.

Simply maximizing the guide-policy with respect to $V(s)$ will result in a state where the execute-policy may make erroneous generalization. To alleviate this issue, we add a behavior constraint term to the learning objective of the guide-policy. Denote the guide-policy as $g_{\omega}(s): \mathcal{S} \rightarrow \mathcal{S}$, the learning objective is given by
\begin{equation}
    \max_{\omega} \mathbb{E}_{s \sim \mathcal{D}, s' \sim g_{\omega}(s)} \Big[V_{\phi}(s') + \alpha \log g_{\mu}(s'|s) \Big],
    \label{eq: update g}
\end{equation}
where $g_{\mu}$ is an estimated behavior guide-policy from the dataset. The weight $\alpha$ serves as the trade-off between guiding to space with high-reward and space that the execute-policy ought to have a correct generalization.
We also give an alternative learning objective that implicitly involves the behavior constraint by using the residual ($r + \gamma V_{\phi'}(s') - V_{\phi}(s)$) as the behavior cloning weight \cite{peters2007reinforcement, peng2019advantage}, which bypasses the need to estimate $g_{\mu}$ by
\begin{equation}
    \max_{\omega} \mathbb{E}_{(s, s') \sim \mathcal{D}} \bigg[\exp \Big(\frac{r + \gamma V_{\phi'}(s') - V_{\phi}(s)}{\alpha} \Big) \log g_{\omega}(s'|s) \bigg].
    \label{eq: update g two}
\end{equation}

\subsection{Learning the Execute-Policy: Training and Evaluation}
\begin{wrapfigure}{R}{0.45\textwidth}
\vspace{-0.7cm}
    \begin{minipage}{0.45\textwidth}
        \begin{algorithm}[H]
            \caption{Policy Guided Offline RL}
            \label{alg:por}
            \begin{algorithmic}[1]
                \Require $\mathcal{D}$, $\tau$, $\alpha$, $g_{\mu}$ (optional).
                \State {\color{blue}{// Training}}
                \State Initialize $V_{\phi}$, $V_{\phi^{\prime}}$, $g_{\omega}$, $\pi_{\theta}$
                \For{$t=1, 2,\cdots, N$}
                \State Sample transitions $(s, r, s') \sim \mathcal{D}$
                \State Update $V_{\phi}$ by Eq.(\ref{eq: update V})
                \State Update $g_{\omega}$ by Eq.(\ref{eq: update g}) or Eq.(\ref{eq: update g two})
                \State Update $V_{\phi^{\prime}}$ by $\phi^{\prime} \leftarrow \lambda \phi + (1-\lambda) \phi'$
                \EndFor
                
                \For{$t=1, 2,\cdots, M$}
                \State Sample transitions $(s, a, s') \sim \mathcal{D}$
                \State Update $\pi_{\theta}$ by Eq.(\ref{eq: execute policy})
                \EndFor
                \State  {\color{blue}{// Evaluation}}
                \State Get initial state $s$, set $d$ as False
                \While {not $d$}
                \State Get action $a$ form Eq.(\ref{eq: evaluation})
                \State Roll out $a$ and get $(s^{\prime}, r, d)$
                \State Set $s=s'$
                \EndWhile
            \end{algorithmic}
        \end{algorithm}
    \end{minipage}
\vspace{-0.1cm}
\end{wrapfigure}

After learning the guide-policy, now we turn to the execute-policy. Since the job of the execute-policy is to have a strong generalization ability, we adopt the \textit{RL via Supervised Learning} framework by conditioning the execute-policy on $s'$ that encountered in the dataset.
% at state $s$.
To be more specific, denote the execute-policy as $\pi_{\theta}(s,s'): \mathcal{S} \times \mathcal{S} \rightarrow [0, 1]$. During training, $\pi$ performs supervised learning by maximizing the likelihood of the actions given the states and next states, yielding the following objective
% training execute-policy 
\begin{equation}
    \max_{\theta} \ \mathbb{E}_{(s, a, s') \in \mathcal{D}}\Big[\log \pi_{\theta}(a | s, s')\Big],
    \label{eq: execute policy}
\end{equation}
in some scenarios with low data quality, one can also add the exponential residual weight in objective (\ref{eq: update g two}) to (\ref{eq: execute policy}) to mitigate the effect of those bad actions.

During evaluation, given a state $s$, the final action is determined by both the guide-policy and the execute-policy, by
\begin{equation}
    a = \arg \max_{a} \ \pi_{\theta}(a | s, g_{\omega}(s)).
    \label{eq: evaluation}
\end{equation}

Note that our learning objective (\ref{eq: execute policy}) differs from previous \textit{RL via Supervised Learning} methods in that we only use the next state $s'$ as the conditioned variable, there's no need to estimate the cumulative return \cite{chen2021decision}, which may be highly suboptimal in certain cases \cite{eysenbach2022imitating,brandfonbrener2022does}. Our method also works in settings where we don't know the goal information \cite{emmons2021rvs}. Also, during evaluation, previous methods show that the choice of conditioned variable is crucial important as little changes will cause significant performance difference \cite{chen2021decision}.
In our method, owing to the existence of the guide-policy, the optimal conditioned variable can be automatically generated.
% \hoyin{add more discussion}

Our final algorithm, \texttt{POR}, consists of three supervised stages: learning $V$, learning $g$, and learning $\pi$. 
We summarize the training and evaluation procedure of \texttt{POR} in Algorithm \ref{alg:por}. 
Note that the training of $g$ and $\pi$ are fully decoupled, this brings \texttt{POR} some nice properties to improve with supplementary data or transfer to new tasks, we will further investigate it in the experiments.

% The concept of generating states is reminiscent of hierarchical RL, in which the policy is implemented as a hierarchy of sub-policies. In particular, approaches related to feudal RL rely on a manager policy providing goals (possibly indirectly, through sub-manager policies) to a worker policy. These goals generally map to actual environment states, either through a learned state representation as in FeUdal Networks , an engineered representation as in h-DQN, or simply by using the same format as raw environment states as in HIRO One could think of the function in QSS as operating like a manager by suggesting a target state, and of the I function as operating like a worker by providing an action that reaches that state. Unlike with hierarchical RL, however, both operate at the same time scale.

\subsection{Analysis} 
% \hoyin{more? deterministic mdp?}
\label{sec: theory}
% Beyond these observations, we are inquisitive to find a theoretical explanation to support our claim. 
In this section, we give a theoretical analysis of \texttt{POR}. Concretely, we aim to 1) give the lower bound of the performance difference between \texttt{POR} and the optimal policy $\pi^{*}$, and 2) analyze how the guide-policy $g$ will influence this bound.

We begin by introducing the notation and assumptions used in our analysis.
We denote $P(s,s'): \mathcal{S}\times\mathcal{S}\rightarrow\mathcal{A}$ as the \textit{inverse transition operator}. We denote $a_g$ and $a$ as the ground truth of $\pi(s,g(s))$ and $\pi(s, s')$, respectively. We also denote $\epsilon:=\sup_{(s,a, s')\in\mathcal{D}}\|\pi(s,s')-a\|$ as the upper bound of the approximation error in the dataset during training. We then introduce the following two assumptions.

\begin{assumption} (Non-lazy MDP)
    The inverse transition operator $P(s,s')$ is $L_1$-Lipschitz continuous, i.e., $\|P(s_1,s_1')-P(s_2,s_2')\|\le L_1\|(s_1,s_1')-(s_2,s_2')\|$.
    \label{assumption:smooth_transition}
\end{assumption}

\begin{assumption} (Lipschitz continuous function approximators)
    The execute-policy we trained is $L_2$-Lipschitz continuous, i.e., $\|\pi(s_1,s_1')-\pi(s_2, s_2')\|\le L_2\|(s_1,s_1')-(s_2,s_2')\|$.
    \label{assumption:smooth_policy}
\end{assumption}

Assumption \ref{assumption:smooth_transition} holds when actions do have effects on state transitions (i.e. if $a\neq0$, then $\|s'-s\|\ge \epsilon_s > 0$). We refer the MDP that satisfies Assumption \ref{assumption:smooth_transition} as \textit{non-lazy MDP} and others as \textit{lazy MDP}. It should be mentioned that most real-world tasks belong to the \textit{non-lazy MDP} case, it is meaningless to study under the \textit{lazy MDP} case because in \textit{lazy MDP}, action changes will have little effect on state transitions, making the policy learning in vain. Assumption \ref{assumption:smooth_policy} is a mild assumption that is frequently utilized in plenty of works \cite{virmaux2018lipschitz, miyato2018spectral, liu2022learning}.

Then in Theorem \ref{theorem:ljx}, we show that under these two assumptions, the gap between the optimal action and the action output by \texttt{POR} can be bounded.

\begin{theorem}
(Single step gap to optimal action). The single-step gap between optimal action and the action induced by our method can be bounded as
    % \begin{equation}
    \begin{align}
        \label{equ:single_step_bound}
        \|\pi(s,g(s))-a^*\|
        % &= \|\pi(s,g(s))-\pi(s,s')+\pi(s,s') - a + a -a_g+a_g-a^*\|\\
        % &\le \|\pi(s,g(s))-\pi(s,s')\| + \|\pi(s,s')-a\| + \|a-a_g\| + \|a_g-a^*\|\ \ {(\rm Triangle)}\\
        % &\le L_2\|g(s)-s'\|+\epsilon+L_1\|g(s)-s'\| +\|a_g-a^*\| \\
        &\le \underbrace{(L_1+L_2)\|g(s)-s'\|}_{l_1}+\underbrace{\|a_g-a^*\|}_{l_2}+\underbrace{\epsilon}_{l_3}
    \end{align}
    % \end{equation}
    \label{theorem:ljx}
\end{theorem}
Theorem \ref{theorem:ljx} states that single step optimal gap is related to three parts: $l_1,l_2,l_3$. $l_1$ is the generalization performance factor. It can be found that a small value of $\|g(s)-s'\|$ enables a small $l_1$, which indicates the necessity to constrain the guide-policy to stay close to the dataset. 
$l_2$ denotes the suboptimality constant. Intuitively, a loosened constraint on the guide-policy may induce a small suboptimality constant but on the other hand, may suffer from the risk of a high value of $l_1$. 
The guide-policy $g$ serves as a trade-off between $l_1$ and $l_2$, we could balance the constraint strength and suboptimality by adjusting $\alpha$ in Eq.(\ref{eq: update g}) to obtain the lowest bound. 
$l_3$ is related to the approximation error on the training data, which can be reduced via improving the representation ability of function approximators but may suffer from a high $L_2$ due to the overfitting caused by over-parameterization. 
$l_1$ also depends on the Lipschitz constant $L_1$ and $L_2$. Note that a smooth \textit{inverse transition operator} will result in a small value of $L_1$ and $L_2$ (because $\pi(s,s')$ is trained to approximate $P(s,s')$), this means our method may achieve better performance under a smoother inverse transition. 

Based on Theorem \ref{theorem:ljx}, we can also give the performance bound of our method by replacing the $\epsilon$ in Theorem 3.2 in \cite{ghosh2020learning} with the RHS of Eq.(\ref{equ:single_step_bound}): $J(\pi^*)-J(\pi)\le \sup_{s,a,h}(l_1+l_2+l_3)T$, where $T$ is the horion of each episode, $h$ is the time step.

\section{Experiments}
We present empirical evaluations of \texttt{POR} in this section. 
We first evaluate \texttt{POR} against other baseline algorithms on D4RL \cite{fu2020d4rl} benchmark datasets.
We then explore deeper on the guide-policy about the benefits of the decoupled training process. 
We finally establish ablation studies on the execute-policy.

\subsection{Comparative Experiments on D4RL Benchmark Datasets}
% introduce the setting, difference gym/mujoco, why maze is important

\begin{table}[tbp]
\centering
\small
\caption{Averaged normalized scores of \texttt{POR} against other baselines. The scores are taken over the final 10 evaluations with 5 seeds. In the \textit{Best RL Baseline} column, the algorithm with the best performance among 5 RL-based algorithms (BCQ \cite{fujimoto2018addressing}, BEAR \cite{kumar2019stabilizing}, BRAC \cite{wu2019behavior}, CQL \cite{kumar2020conservative}, TD3+BC \cite{fujimoto2021minimalist}) is presented. \texttt{POR} achieved the highest scores in 11 out of 18 tasks. 
} \label{table:results}
% \renewcommand\arraystretch{1.5}
% \vspace{5pt} 
\resizebox{0.95\linewidth}{!}{
\begin{tabular}{lccccccc} 
\toprule
\multirow{2}{*}{D4RL Dataset}                               & \multicolumn{3}{c}{\begin{tabular}[c]{@{}c@{}}Weighted BC\\\end{tabular}} & \multicolumn{3}{c}{\begin{tabular}[c]{@{}c@{}}Conditioned BC\\\end{tabular}}                                     & \multirow{2}{*}{\begin{tabular}[c]{@{}c@{}}Best RL \\Baseline\end{tabular}}  \\ 
\cmidrule(r){2-4}\cmidrule(r){5-7}
                                                       & 10\%BC                                        & One-step  & IQL          & DT        & RvS                                             & POR (ours)                                         &                                                                           \\ 
\hline
antmaze-u                                              & 62.8                                           & 64.3      & 87.5{\color[HTML]{525252} $\pm$2.6}     & 59.2      & 65.4                                            & \begin{tabular}[c]{@{}c@{}}\colorbox{mine}{90.6}{\color[HTML]{525252} $\pm$7.1}\\\end{tabular} & $89.0^{\text{BCQ}}$                                                           \\
antmaze-u-d                                            & 50.2                                           & 60.7      & 66.2{\color[HTML]{525252} $\pm$13.8}    & 53.0      & 60.9                                            & \colorbox{mine}{71.3}{\color[HTML]{525252} $\pm$12.1}                                          & $61.0^{\text{BEAR}}$                                                             \\ 
\hline
antmaze-m-p                                            & 5.4                                            & 0.3       & 71.2{\color[HTML]{525252} $\pm$7.3}     & 0         & 58.1                                            & \colorbox{mine}{84.6}{\color[HTML]{525252} $\pm$5.6}                                           & $68.0^{\text{CQL}}$                                                              \\
\begin{tabular}[c]{@{}l@{}}antmaze-m-d\\\end{tabular}  & 9.8                                            & 0         & 70.0{\color[HTML]{525252} $\pm$10.9}    & 0         & 67.3                                            & \colorbox{mine}{79.2}{\color[HTML]{525252} $\pm$3.1}                                          & $68.0^{\text{CQL}}$                                                              \\ 
\hline
antmaze-l-p                                            & 0                                              & 0         & 39.6{\color[HTML]{525252} $\pm$5.8}     & 0         & 32.4                                            & \colorbox{mine}{58.0}{\color[HTML]{525252} $\pm$12.4}                                         & $18.8^{\text{CQL}}$                                                     \\
antmaze-l-d                                            & 0                                              & 0         & 47.5{\color[HTML]{525252} $\pm$9.5}     & 0         & 36.9                                            & \colorbox{mine}{73.4}{\color[HTML]{525252} $\pm$8.5}                                           & $45.6^{\text{CQL}}$                                                            \\ 
\hline
antmaze mean                                          & 21.4                                          & 20.8     &      63.6 {\color[HTML]{525252} $\pm$8.3}       & 18.7     & 53.5                                             &       \colorbox{mine}{76.2} {\color[HTML]{525252} $\pm$8.1}                                             &          58.4                                                                 \\ 
\hline
% \hline
% halfcheetah-r                                          & 5.4                                            & 3.7{\color[HTML]{525252} $\pm$0.2}   &       11.2{\color[HTML]{525252} $\pm$2.9}      &    -       &     -                                            & \colorbox{mine}{29}{\color[HTML]{525252} $\pm$0.7}                                             &    $20.0^{\text{CQL}}  $                                                                        \\
% hopper-r                                               & 4.2                                            & 5.2{\color[HTML]{525252} $\pm$0.2}   &        7.9{\color[HTML]{525252} $\pm$0.4}      &     -      &       -                                          & 12{\color[HTML]{525252} $\pm$2.1}                                             &    \colorbox{mine}{$14.2$}$^{\text{BEAR}}$                                                                        \\
% walker2d-r                                             & 6.7                                        & 5.6{\color[HTML]{525252} $\pm$0.6}   &        5.9{\color[HTML]{525252} $\pm$0.5}     &     -      &      -                                           & 6.3{\color[HTML]{525252} $\pm$0.3}                                           &   \colorbox{mine}{$8.3$}$^{\text{CQL}}  $                                                                         \\ 
\hline
halfcheetah-m                                          & \begin{tabular}[c]{@{}c@{}}42.5\\\end{tabular} & 48.4{\color[HTML]{525252} $\pm$0.1}  & 47.4{\color[HTML]{525252} $\pm$0.2}     & 42.6{\color[HTML]{525252} $\pm$0.1}  & 41.6                                            &  \colorbox{mine}{48.8}{\color[HTML]{525252} $\pm$0.5}                                           &     $48.3^{\text{TD3+BC}}  $                                                                    \\
hopper-m                                               & 56.9                                           & 59.6{\color[HTML]{525252} $\pm$2.5}  & 66.2{\color[HTML]{525252} $\pm$5.7}    & 67.6{\color[HTML]{525252} $\pm$1.0}  & 60.2                                            &  \colorbox{mine}{78.6}{\color[HTML]{525252} $\pm$7.2}                                           &   $59.3^{\text{TD3+BC}}  $                                                                      \\
walker2d-m                                             & 75.0                                           & 81.8{\color[HTML]{525252} $\pm$2.2}  & 78.3{\color[HTML]{525252} $\pm$8.7}     & 74.0{\color[HTML]{525252} $\pm$1.4}  & 71.7                                            & 81.1{\color[HTML]{525252} $\pm$2.3}                                           &  \colorbox{mine}{$83.7$}$^{\text{TD3+BC}} $                                                                 \\ 
\hline
halfcheetah-m-r                                       & 40.6                                           & 38.1{\color[HTML]{525252} $\pm$1.3}  & 44.2{\color[HTML]{525252} $\pm$1.2}     & 36.6{\color[HTML]{525252} $\pm$0.8}  & 38.0                                            & 43.5{\color[HTML]{525252} $\pm$0.9}                                           &    \colorbox{mine}{$45.5$}$^{\text{CQL}} $                    \\
hopper-m-r                                         & 75.9                                           & 97.5{\color[HTML]{525252} $\pm$0.7}  & 94.7{\color[HTML]{525252} $\pm$8.6}     & 82.7{\color[HTML]{525252} $\pm$7.0} & 73.5                                            & \colorbox{mine}{98.9}{\color[HTML]{525252} $\pm$2.1}                                           &        $95.0^{\text{CQL}} $                                                                   \\
walker2d-m-r                                          & 62.5                                           & 49.5{\color[HTML]{525252} $\pm$12.0}   & 73.8{\color[HTML]{525252} $\pm$7.1}    & 66.6{\color[HTML]{525252} $\pm$3.0}  & 60.6                                            & 76.6{\color[HTML]{525252} $\pm$6.9}                                           &   \colorbox{mine}{$81.8$}$^{\text{TD3+BC}}$                                                                        \\ 
\hline
halfcheetah-m-e                                        & 92.9                                           & 93.4{\color[HTML]{525252} $\pm$1.6}  & 86.7{\color[HTML]{525252} $\pm$5.3}     & 86.8{\color[HTML]{525252} $\pm$1.3}  & 92.2                                            & \colorbox{mine}{94.7}{\color[HTML]{525252} $\pm$2.2}                                           &    $91.6^{\text{CQL}} $                                                                   \\
hopper-m-e                                             & 110.9                                          & 103.3{\color[HTML]{525252} $\pm$1.9} & 91.5{\color[HTML]{525252} $\pm$14.3}    & \colorbox{mine}{107.6}{\color[HTML]{525252} $\pm$1.8} & \begin{tabular}[c]{@{}c@{}}101.7\\\end{tabular} & 90.0{\color[HTML]{525252} $\pm$12.1}                                          & $105.4^{\text{CQL}}$                                                                    \\
\begin{tabular}[c]{@{}l@{}}walker2d-m-e\\\end{tabular} & 109.0                                          & \colorbox{mine}{113.0}{\color[HTML]{525252} $\pm$0.4}   & 109.6{\color[HTML]{525252} $\pm$1.0}    & 108.1{\color[HTML]{525252} $\pm$0.2} & 106.0                                           & 109.1{\color[HTML]{525252} $\pm$0.7}                                          &       $110.1^{\text{TD3+BC}} $                                                                   \\ 
\hline
locomotion mean                                       & 55.5                                          &     58.2  {\color[HTML]{525252} $\pm$2.0}     &   57.6  {\color[HTML]{525252} $\pm$6.6}           &   56  {\color[HTML]{525252} $\pm$1.4}         & 53.7                                           &   \colorbox{mine}{65.6}  {\color[HTML]{525252} $\pm$2.7}                                                  &    63.5                                                               \\
\bottomrule
\end{tabular}}
\end{table}

\begin{table}[tbp]
\centering
\small
\caption{Performance of \texttt{POR} against \texttt{POR-sparse} on challenging AntMaze datasets.} \label{table:results_sql}
% \vspace{5pt}
\resizebox{0.9\linewidth}{!}{
\begin{tabular}{lcccccc} 
\toprule
Algorithm    & antmaze-u    & antmaze-u-d    & antmaze-m-p    & antmaze-m-d    & antmaze-l-p    & antmaze-l-d    \\
\hline
POR          & 90.6 {\color[HTML]{525252} $\pm$10.7}    & 71.3 {\color[HTML]{525252} $\pm$12.2}    & \colorbox{mine}{84.6} {\color[HTML]{525252} $\pm$5.6}    & \colorbox{mine}{79.2} {\color[HTML]{525252} $\pm$3.1}    & \colorbox{mine}{58.0} {\color[HTML]{525252} $\pm$12.4}    & \colorbox{mine}{73.4} {\color[HTML]{525252} $\pm$8.5}    \\
POR-Sparse      & \colorbox{mine}{95.0} {\color[HTML]{525252} $\pm$3.4}    & \colorbox{mine}{86.3} {\color[HTML]{525252} $\pm$6.1}    & 82.4 {\color[HTML]{525252} $\pm$5.5}    & 71.0 {\color[HTML]{525252} $\pm$2.4}    & 48.8 {\color[HTML]{525252} $\pm$4.1}    & 53.4 {\color[HTML]{525252} $\pm$3.6}    \\
\bottomrule
\end{tabular}}
\end{table}

We first evaluate our approach on D4RL MuJoCo and AntMaze datasets \cite{fu2020d4rl}. 
Notice that most of MuJoCo datasets (except \texttt{random} and \texttt{medium} datasets) consist of a large fraction of near-optimal trajectories, while AntMaze datasets contain few or no near-optimal trajectories. Hence those low-quality datasets can serve as a good testbed for quantifying the \textit{compositionality} ability. 

We compare \texttt{POR} with both RL-based and imitation-based baselines. 
For RL-based baselines, we select the best one from 5 state-of-the-art algorithms, including BCQ \cite{fujimoto2018addressing}, BEAR \cite{kumar2019stabilizing}, BRAC \cite{wu2019behavior}, CQL \cite{kumar2020conservative} and TD3+BC \cite{fujimoto2021minimalist}.
For imitation-based baselines, we further split them into \textit{Weighted BC} methods and \textit{Conditioned BC} methods.
Weighted BC methods include 10\%BC \cite{chen2021decision}, One-step RL \cite{brandfonbrener2021offline} and IQL \cite{kostrikov2021offline}.
10\%BC is a filtered version of BC that runs behavior cloning on only the top 10\% high-return trajectories in the dataset. One-step RL and IQL can be viewed as using different weights to do behavior cloning (see Appendix \ref{sec: discuss} for discussion). 
Conditioned BC methods include DT \cite{chen2021decision} and RvS \cite{emmons2021rvs}. DT is built with Transformer and conditioned on the cumulative return. RvS well studied the effect of different conditional information on different tasks. The score of RvS is chosen by the higher score between RvS-R and Rvs-G, which use the reward-to-go and \textbf{oracle} goal information as the conditioned variable, respectively.
Note that we also categorize our method \texttt{POR} into conditioned BC methods because the execute-policy is learned in a similar manner to other conditioned BC methods.
Note that the results of Filter BC are from our own implementation, and the results of IQL on MuJoCo random tasks are re-runed using the author-provided implementation.
% \footnote{\url{https://github.com/ikostrikov/implicit_q_learning}}.
Other results are taken directly from their corresponding papers. 
To enable fairness and consistency in the evaluation process, we train our algorithm for 1 million time steps and evaluate every 5000 time steps that consist of 10 episodes. Full experimental details are included in Appendix \ref{appendix: exp d4rl}, learning curves of \texttt{POR} against our implemented \texttt{IQL} (using the same codebase and network structures) and other ablation studies can be found in Appendix \ref{appendix: learning curve} and Appendix \ref{appendix: ablation study}, respectively. 

The results are shown in Table \ref{table:results}, in MuJoCo locomotion tasks, \texttt{POR} performs competitively to the best performance of prior methods in high-quality datasets, while achieving much better performance than other methods in low-quality datasets (e.g., \texttt{medium} datasets). 
On the more challenging AntMaze tasks, \texttt{POR} outperforms all other baselines by a large margin. It even surpassed RvS that uses the oracle goal information to learn the policy.
We conjecture that the success in all those datasets should be credited to the out-of-distribution generalization ability of \texttt{POR}.
We also compare \texttt{POR-sparse} with \texttt{POR}, in some AntMaze tasks, the performance of \texttt{POR-sparse} even surpasses that of \texttt{POR}.

\begin{figure}[tbp]
\resizebox{\linewidth}{!}{
		\centering
		\subfigure[antmaze-large-diverse]{
			\begin{minipage}[b]{0.6\textwidth}
				\includegraphics[width=1.0\textwidth]{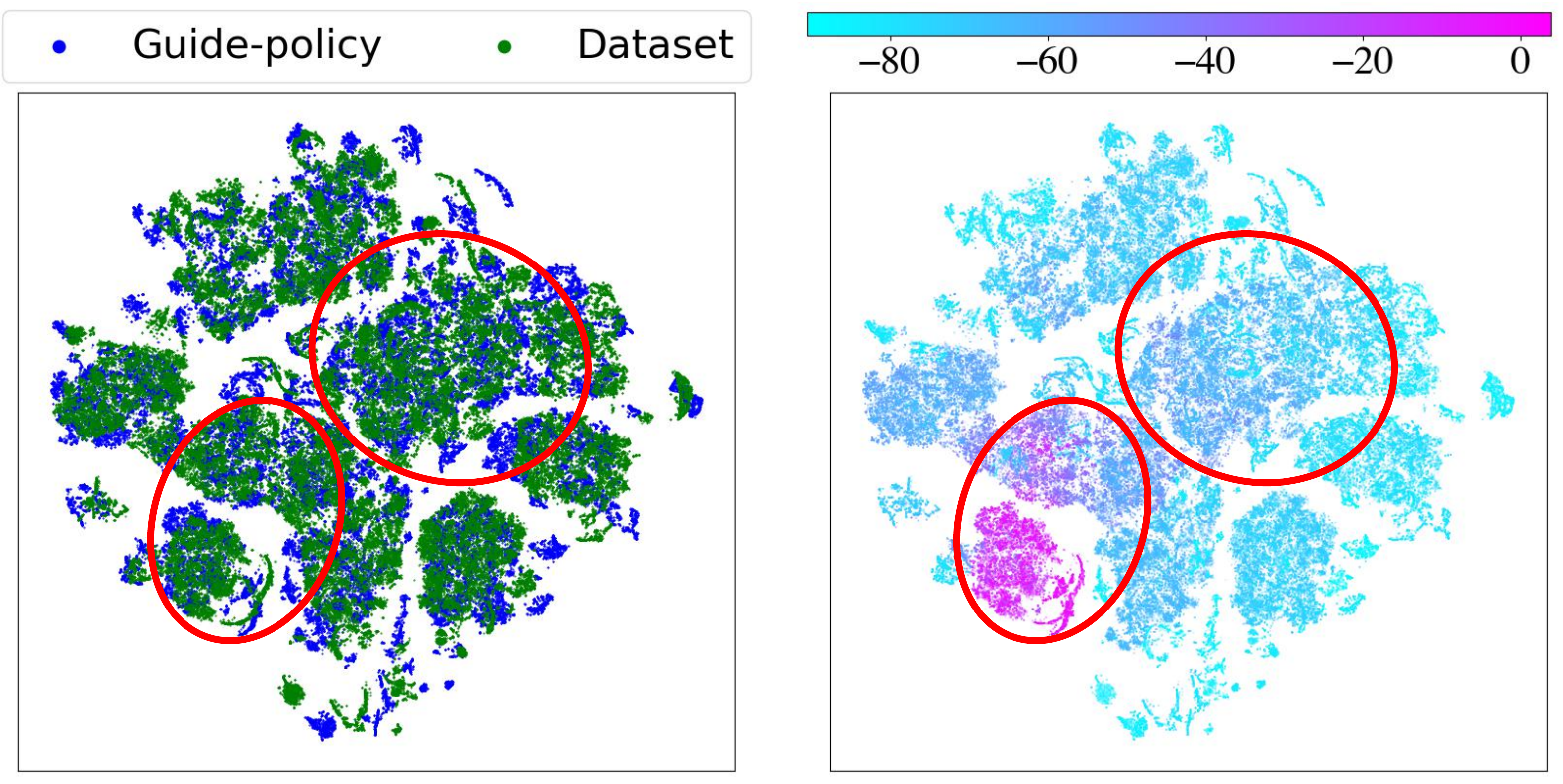}
			\end{minipage}
		}
		\subfigure[hopper-medium-replay]{
			\begin{minipage}[b]{0.6\textwidth}
				\includegraphics[width=1.0\textwidth]{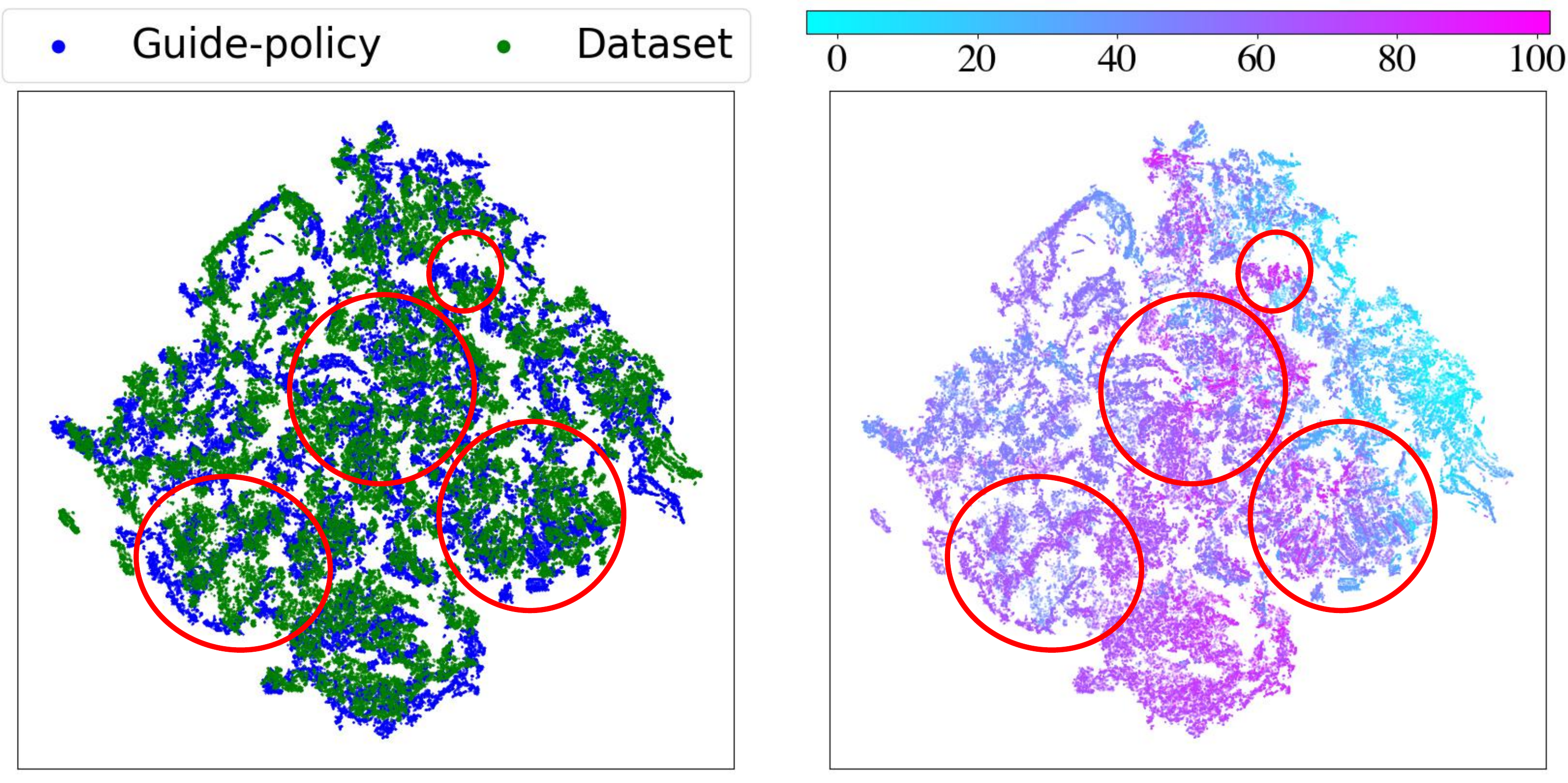}
			\end{minipage}
		}}
\caption{Visualizations of the distribution of states generated by the guide-policy and states in the dataset, on \texttt{antmaze-large-diverse} and \texttt{hopper-medium-replay} datasets.
% Data points distribution visualization and their corresponding true value, as well as from the corresponding guide-policy in \texttt{antmaze-large-diverse} and \texttt{hopper-medium-replay} tasks. 
The left panel of each subfigure shows the $100,000$ data points of $g(s)$ ({\color{Blue} blue}) and $s^{\prime}$ ({\color{ForestGreen} green}). The right panel of each subfigure shows the corresponding value of each state. The distribution of $g(s)$ and $s'$ bear some similarity but $g(s)$ is more prone to generate high-value states, especially on \texttt{antmaze-large-diverse} dataset.
Better zoom in to see a more clear comparison.
}
\label{fig: state visualization}
\end{figure}

\subsection{Validation Experiments on Out-of-Distribution Generalization} 
% The results discussed in the previous section suggest that CPQ outperforms other baselines on several challenging tasks. 
In this section, we try to validate our hypothesis about the out-of-distribution generalization of \texttt{POR}.
We are especially interested to see how the guide-policy is learned in those high-dimensional tasks in D4RL.
To do so, we compare the distribution of $g(s)$ generated by the guide-policy with the distribution of $s^{\prime}$ from the offline dataset. 
We choose \texttt{antmaze-large-diverse} and \texttt{hopper-medium-replay}, two datasets for evaluation.
% as those two are the most representative datasets where \texttt{POR} is much better than prior methods. 
For each dataset, we randomly collect $100,000$ $(s, s^{\prime})$ transition tuples. 
To visualize the results clearly, we plot the distribution of $g(s)$ and $s^{\prime}$ with t-Distributed Stochastic Neighbor Embedding (t-SNE) \cite{hinton2002stochastic}, we also visualize the true values of all states using the optimal value function $V^{*}(s)$.
% we plot the distribution of $s'$ with green dots, $g(s)$ generated from the guide-policy with blue dots, and their corssponding true value from the true value function . 
% \footnote{We choose the hyperparamters of perplexity with 50, n components with 2, init with pca and iterations with 5,000.}. 

The results are shown in Figure \ref{fig: state visualization}, it can be found that in both tasks, the distribution of $g(s)$ bears some similarity to the distribution of $s^{\prime}$. However, when looking at the non-overlapped area of $g(s)$ and $s^{\prime}$, we found that $g(s)$ has a much higher value than $s^{\prime}$, especially in the \texttt{antmaze-large-diverse} dataset. 
% The disagreement of the distribution brings the advantage of higher values.
Those out-of-distribution and high-value areas are exactly where \texttt{POR} generalized to, note that the high-value benefit of the single-step generalization could also be accumulated and propagated through time steps, resulting in a much better policy.

\subsection{Investigations on the Guide-Policy} \label{exp: guide-policy} 
In this section, we investigate what is the potential benefit of decoupling the training process of the guide-policy and execute-policy. We found that as the execute-policy is not task-specific and only cares about the effect of actions on different states, one can reuse its powerful generalization ability when encountered with additional suboptimal data or adapting to a new task. We only need to re-train the guide-policy as it is task-specific.

\noindent \textbf{Improve guide-policy by additional suboptimal data} \quad  
We first study the setting where we already have a policy learned from a small yet high-quality dataset $\mathcal{D}_{e}$, and now we get a supplementary large dataset $\mathcal{D}_{o}$. $\mathcal{D}_{o}$ may be sampled from one or multiple behavior policies, it provides higher data coverage but could be suboptimal. 
% The policy learned by $\mathcal{D}_{e}$ is suboptimal due to poor generalization caused by limited data size, and 
We are interested to see whether we can use $\mathcal{D}_{o}$ to obtain a better policy.
This setting is realistic as low-quality datasets appear more often in real-world tasks \cite{yu2022leverage,xu2022discriminator}.
% We first study the offline learning setting where the agent has only access to a small high-quality dataset $\mathcal{D}_{e}$, but also has a large offline dataset $\mathcal{D}_{o}$ sampled from one or multiple behavior policies that could be highly sub-optimal.

\begin{figure}[tbp]
    \centering
    \includegraphics[width=0.8\textwidth]{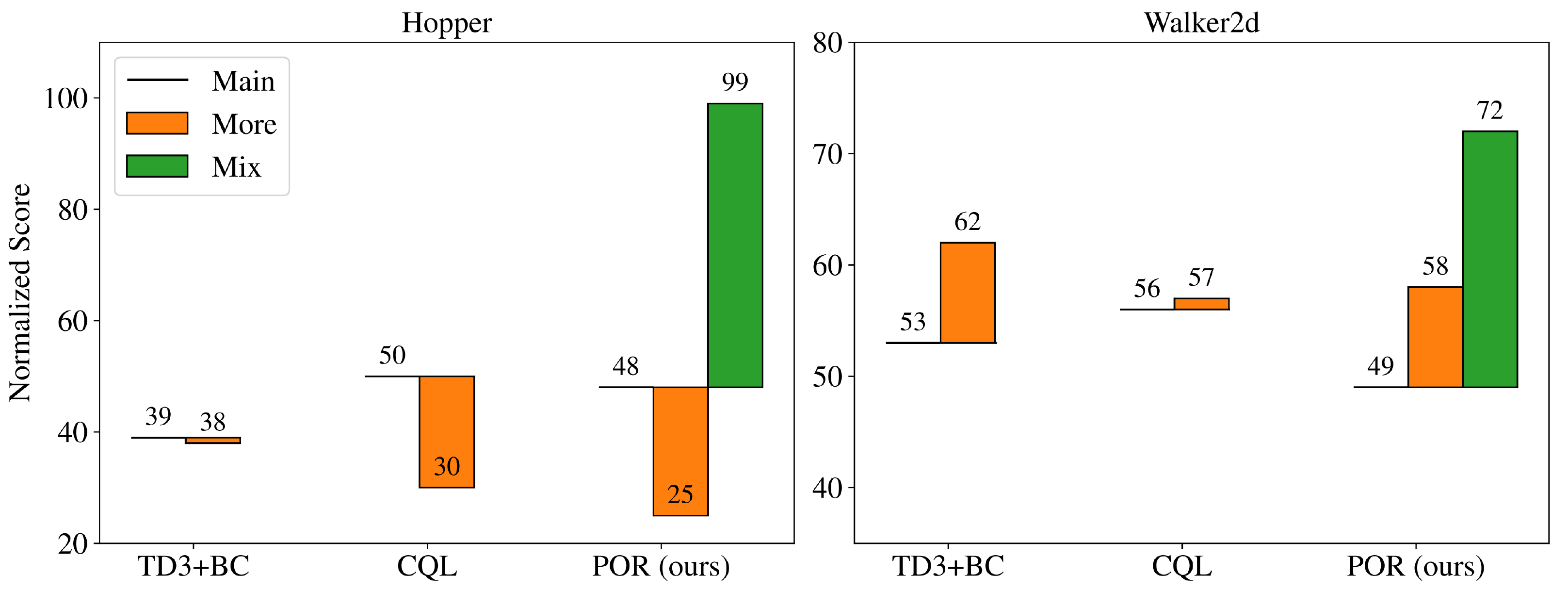}
    \caption{Normalize scores of different algorithms with different training schemes. While the "main" and "more" training schemes can be applied to all methods, the "mix" training scheme is only available to \texttt{POR} because we decouple the training process of $g$ and $\pi$. With the "mix" training scheme, \texttt{POR} outperforms all other algorithms by a large margin, in both \texttt{hopper} and \texttt{walker2d} environments.}
    \label{fig:add data}
\end{figure}

% For the real-world applications with offline RL, the dataset for training is highly probably sub-optimal as well as small amount of additional noise data, which is basically impossible to train alone. 
Na\"ively, we can combine $\mathcal{D}_{e}$ and $\mathcal{D}_{o}$ as a new offline dataset $\mathcal{D}$, then run any off-the-shelf offline RL algorithm. We label this training scheme as "more". We also label the original training scheme that only uses $\mathcal{D}_{e}$ as "main". We test these two training schemes on two state-of-the-art offline RL algorithms, TD3+BC and CQL. The results shown in Figure \ref{fig:add data} indicate that both "main" and "more" are insufficient to learn a good policy, 
The reason why "main" fails is that the resulting policy may not be able to generalize due to the limited size.
While "more" can generalize better than "main" due to access to a much larger dataset, the policy will be negatively impacted by the low-quality data in $\mathcal{D}_{o}$ as both TD3+BC and CQL contain the behavior regularization term.

% We label the training procedure using only $\mathcal{D}_{e}$ as "main", we also label the combination of sub-optimal dataset and additional noise dataset as "more". 
Due to the availability of decoupled training, \texttt{POR} is able to use a different training scheme, we can keep the execute-policy learned from $\mathcal{D}_{e}$ unchanged but use the combination of $\mathcal{D}_{e}$ and $\mathcal{D}_{o}$ to re-train the guide-policy. 
Note that in this scheme, we allow the use of an \textit{action-free} dataset $\mathcal{D}_{o}$, which often appears in real-world scenarios (i.e., video data and third-person demonstrations) \cite{torabi2018behavioral}.
The value function will be more accurate and better generalized when combining $\mathcal{D}_{e}$ with $\mathcal{D}_{o}$ to have a larger data coverage (see \textit{exploration} in online RL as an example).
We label this unique training procedure as "mix". 
It can be shown in Figure \ref{fig:add data} that, by doing so, \texttt{POR} could have a super large improvement over "main" and "more".
The version "mix" brings \texttt{POR} up to $99$, close to the upper limit of the performance in the \texttt{hopper} environment, with only limited data. In the \texttt{walker2d} environment, \texttt{POR} with "mix" also outperforms all other baselines.

\noindent \textbf{Change guide-policy in new tasks} \quad
Then, we study whether we can use the guide-policy to do a lightweight adaptation to new tasks, without changing the execute-policy.
% evaluate the adaptation ability of the guide-policy in new tasks as well as the generalization ability of pretrained execute-policy. As shown in Figure \ref{fig: newtasks}, similar to Trajectory Transfromer \cite{janner2021offline}, we introduce three continuous variants of the classic four rooms environment \cite{sutton1999between}. 
We conduct experiments on the continuous variant of the classic four-room environment \cite{janner2021offline}. The training data consists of trajectories collected by a goal-reaching controller with the start and end locations sampled randomly at non-wall locations, mixed with data collected by a random policy.
% In all three tasks, our training data is generated by a controller from the start state to the goal state, where the start state and the goal state are uniformly sampled from the whole state space. For the evalation process, the start and goal state are uniformly sampled from a specific allowed state space (green box in the left down for the start state and red box in the right up for the goal state). 
In this environment, we introduce three different tasks: Task A (\texttt{Four-room}) needs the agent to travel from the start location to the goal location, besides that, Task B (\texttt{Four-room-river}) and task C (\texttt{Four-room-key}) require the agent to bypass the river or get a key before arriving at the goal location. 
% We include full environmental and experimental details in Appendix \ref{appendix: exp fou room}. 

\begin{figure}[tbp]
\centering
\resizebox{0.9\linewidth}{!}{
		\centering
		\subfigure[Four-room]{
			\begin{minipage}[b]{0.3\textwidth}
				\includegraphics[width=1\textwidth]{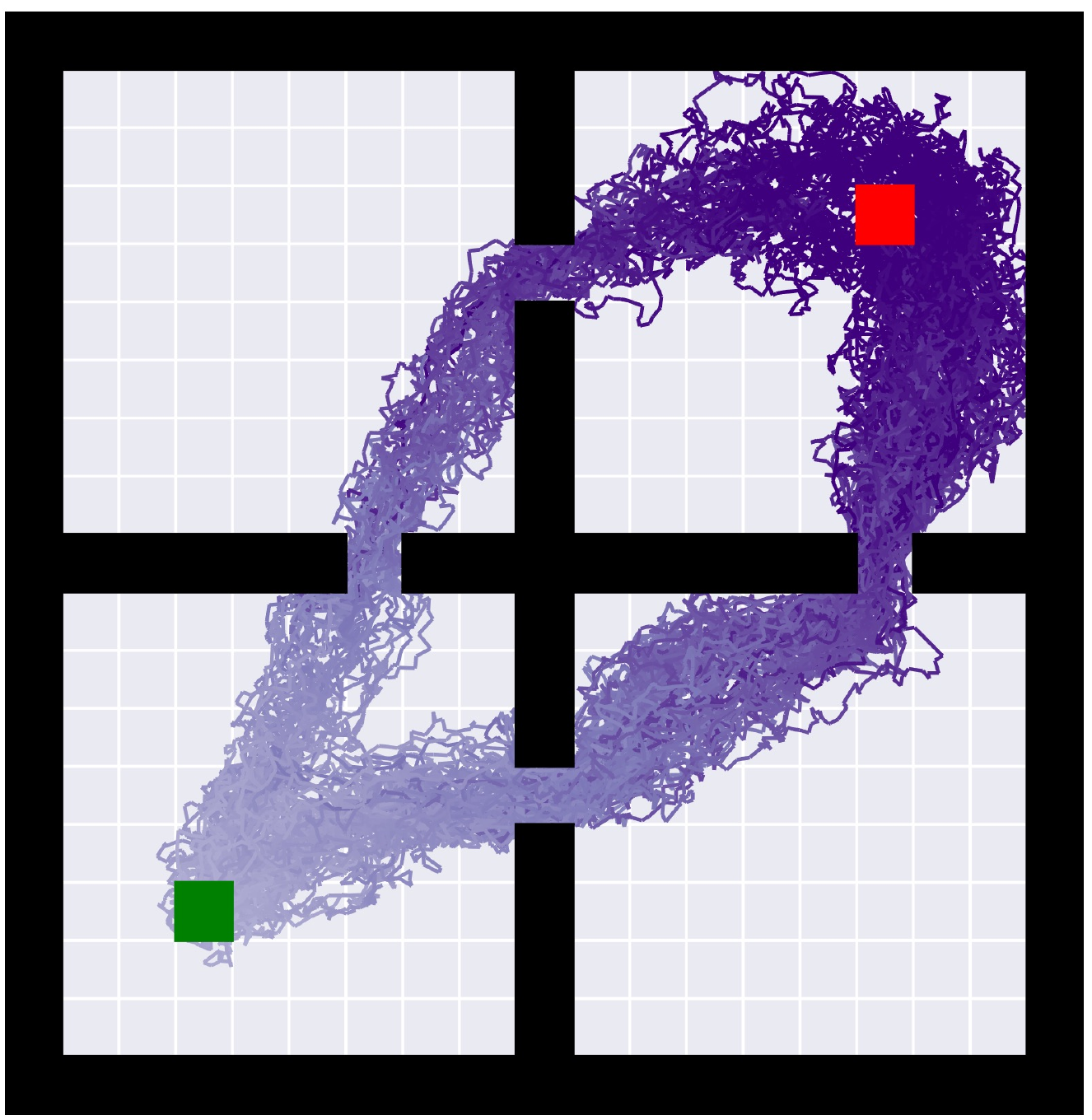}
			\end{minipage}
		}
		\subfigure[Four-room-river]{
			\begin{minipage}[b]{0.3\textwidth}
				\includegraphics[width=1\textwidth]{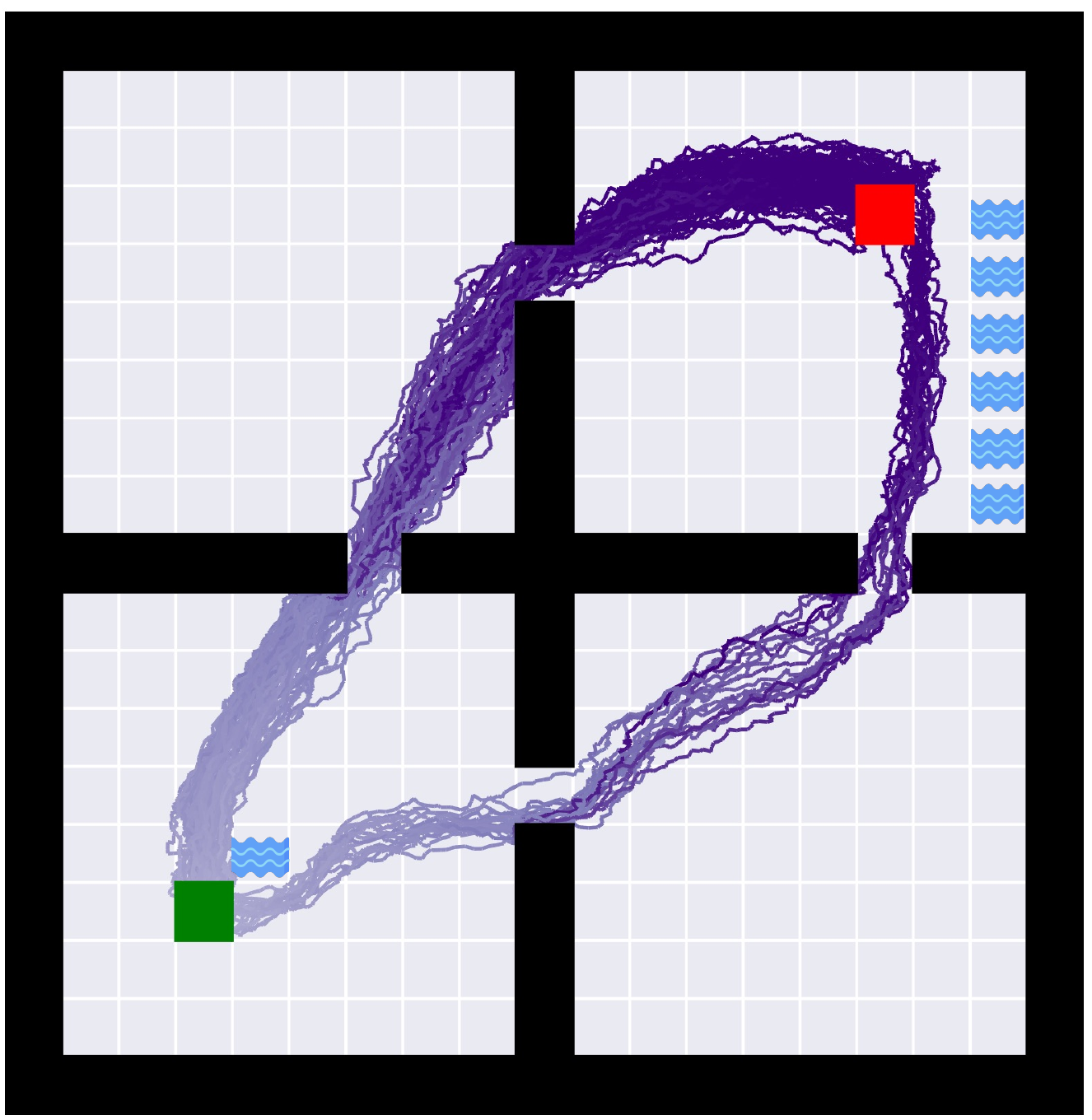}
			\end{minipage}
		}
		\subfigure[Four-room-key]{
			\begin{minipage}[b]{0.3\textwidth}
				\includegraphics[width=1\textwidth]{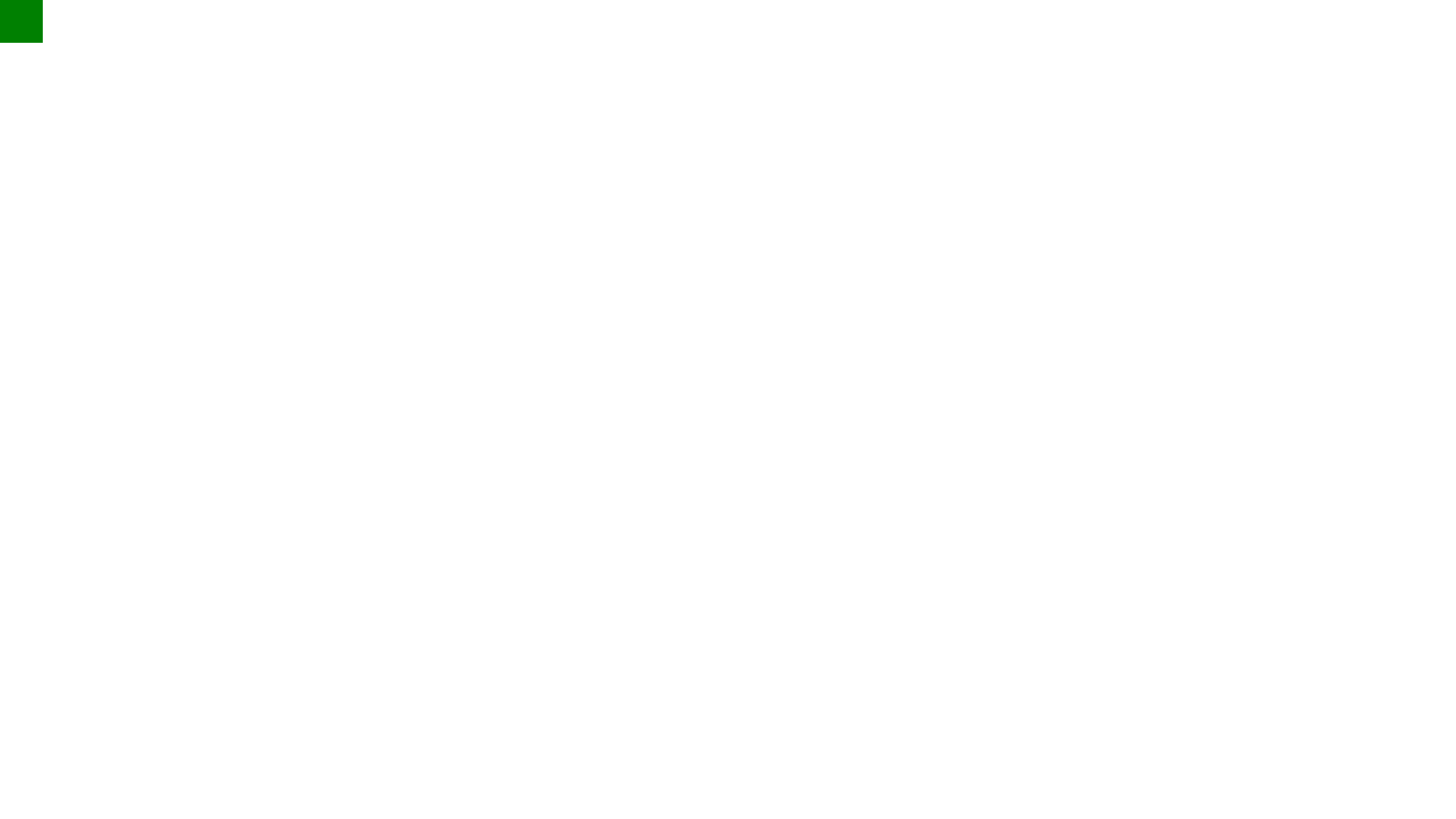}
			\end{minipage}
		}}

\caption{Rollout trajectories of \texttt{POR} with the same pre-trained execute-policy but different guide policies in different four-room tasks, with color becoming more saturated as time progresses. The execute-policy is pre-trained in task A and remains unchanged in task B and task C. All three tasks require the agent to find a path from the start \protect{\includegraphics[height=.3cm]{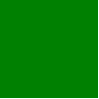}} to the goal \protect{\includegraphics[height=.3cm]{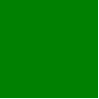}}. Besides, task B requires the agent not to fall into the river \protect{\includegraphics[height=.3cm]{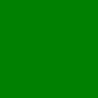}} and task C requires the agent to get the key \protect{\includegraphics[height=.3cm]{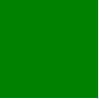}} as well as avoiding the river \protect{\includegraphics[height=.3cm]{Fig/four_room/river.pdf}} before arriving at the goal location. }
\label{fig: newtasks}
\end{figure}

We first train the guide-policy and the execute-policy $\pi$ to solve task A. In task B and task C, we re-train the guide-policy using the corresponding reward function in that task and reuse the execute-policy $\pi$ from Task A. 
% and coupled with the execute-policy $\pi$ for evaluation. 
In Figure \ref{fig: newtasks}, we can see that by doing so, \texttt{POR} is able to successfully solve task B and task C. In detail, in task B, \texttt{POR} is aware of the river and correctly bypasses it. In task C, \texttt{POR} only chooses to pass the bottom door, this is owing to the awareness of the importance of obtaining the key before reaching the goal. This result shows the strong adaptation ablity of the guide-policy for different tasks, as well as the generalization ability of the execute-policy across different tasks.

\section{Conclusions and Limitations}
\label{sec: conclusion and limitation}
In this work, we propose a new offline RL paradigm, \texttt{POR}, which leverages the training stability of imitation-style methods while still encouraging logical out-of-distribution generalization. \texttt{POR} allows \textit{state-compositionality} rather than \textit{action-compositionality} from the dataset. Through theoretical analysis and extensive experiments, we show that \texttt{POR} outperforms prior methods on a variety of datasets, especially those with low quality. We also empirically demonstrate the additional benefits of \texttt{POR} in terms of improving with supplementary suboptimal data and easily adapting to new tasks. 
We hope that \texttt{POR} could shed light on how to enable state-compositionality in offline RL, which connects well to goal-conditioned RL and hierarchical RL \cite{florensa2018automatic}.
One limitation of \texttt{POR} is that the prediction error of the guide-policy may be large when the state space is high-dimensional. However, this can be alleviated with the help of representation learning \cite{laskin_srinivas2020curl, zhang2020learning}.

\noindent \textbf{Acknowledgments} \quad
This work is supported by fundings from AsiaInfo. We thank Michael Janner for providing the code of the continuous four-room environment. We thank Junwu Xiong, James Zhang for their feedback on earlier versions of the manuscript.

\bibliographystyle{plain}
\bibliography{references}
% \bibliographystyle{neurips_2022}

%%%%%%%%%%%%%%%%%%%%%%%%%%%%%%%%%%%%%%%%%%%%%%%%%%%%%%%%%
\newpage
\section*{Checklist}

\begin{enumerate}

\item For all authors...
\begin{enumerate}
  \item Do the main claims made in the abstract and introduction accurately reflect the paper's contributions and scope?
    \answerYes{}
  \item Did you describe the limitations of your work?
    \answerYes{See Section \ref{sec: conclusion and limitation}.} 
  \item Did you discuss any potential negative societal impacts of your work? 
    \answerYes{See Section \ref{sec: conclusion and limitation}.} 
  \item Have you read the ethics review guidelines and ensured that your paper conforms to them?
    \answerYes{}
\end{enumerate}

\item If you are including theoretical results...
\begin{enumerate}
  \item Did you state the full set of assumptions of all theoretical results?
    \answerYes{See Section \ref{sec: theory}.}
        \item Did you include complete proofs of all theoretical results?
    \answerYes{See Appendix \ref{sec: proof}.}
\end{enumerate}

\item If you ran experiments...
\begin{enumerate}
  \item Did you include the code, data, and instructions needed to reproduce the main experimental results (either in the supplemental material or as a URL)?
    \answerYes{See the supplement material.}
  \item Did you specify all the training details (e.g., data splits, hyperparameters, how they were chosen)?
    \answerYes{See Appendix \ref{appendix: exp}.}
        \item Did you report error bars (e.g., with respect to the random seed after running experiments multiple times)?
    \answerYes{In all related figures.}
        \item Did you include the total amount of computing and the type of resources used (e.g., type of GPUs, internal cluster, or cloud provider)?
    \answerYes{See Appendix \ref{appendix: exp}.}
\end{enumerate}

\item If you are using existing assets (e.g., code, data, models) or curating/releasing new assets...
\begin{enumerate}
  \item If your work uses existing assets, did you cite the creators?
    \answerYes{Data is from \cite{fu2020d4rl}.}
  \item Did you mention the license of the assets?
    \answerYes{The license is Apache 2.0.}
  \item Did you include any new assets either in the supplemental material or as a URL?
    \answerYes{See supplement for code and four room dataset.}
  \item Did you discuss whether and how consent was obtained from people whose data you're using/curating?
    \answerNA{Data is simulated.}
  \item Did you discuss whether the data you are using/curating contains personally identifiable information or offensive content?
    \answerNA{Data is simulated.}
\end{enumerate}

\item If you used crowdsourcing or conducted research with human subjects...
\begin{enumerate}
  \item Did you include the full text of instructions given to participants and screenshots, if applicable?
    \answerNA{}
  \item Did you describe any potential participant risks, with links to Institutional Review Board (IRB) approvals, if applicable?
    \answerNA{}
  \item Did you include the estimated hourly wage paid to participants and the total amount spent on participant compensation?
    \answerNA{}
\end{enumerate}

\end{enumerate}

%%%%%%%%%%%%%%%%%%%%%%%%%%%%%%%%%%%%%%%%%%%%%%%%%%%%%%%%%%%%
\clearpage

\appendix

% \section{Broader Impact}
% The aim behind this work is to develop simple yet strong offline RL algorithms, and we believe that \texttt{POR} makes an important step in that direction. We believe that a strong offline RL algorithm, coupled with highly expressive and powerful deep neural networks, will provide us the ability successfully apply end-to-end learning based approaches to a number of real-world problems, such as autonomous driving, robotics, and recommender systems. This will provide considerable societal benefits.

\section{More Discussion} 
\label{sec: discuss}
\noindent \textbf{Why One-step and IQL are imitation-based methods?} \quad 
The core difference between RL-based and imitation-based methods is that RL-based methods learn a value function of policy $\pi$ while imitation-based methods don’t. Learning the value function of $\pi$ requires off-policy evaluation of $\pi$ (i.e., learning $Q^{\pi}$ or $V^{\pi}$), which is prone to distribution shift. The policy evaluation and policy improvement will also affect each other as they are coupled.

Imitation-based methods don't learn $Q^{\pi}$ or $V^{\pi}$, but some of them do learn a value function. We call these methods imitation-based because they learn the value function using only dataset samples, the value function actually tells how advantageous it could be under the behavior policy, much like imitation learning. Also, the policy learning objective of One-step (with exponentially-weighted improvement operator) and IQL can be written as $L(\pi)=\mathbb{E}_{(s, a) \sim \mathcal{D}}\left[\exp \left(\beta\left(Q(s, a)-V(s)\right)\right) \log \pi(a|s)\right]$, which uses dataset actions to doing behavior cloning with different weights (One-step and IQL learn different value functions). 

Cloning dataset actions can only do action-stitching, which loses the ability to surpass the dataset by out-of-distribution (action) generalization. For example, in our toy example in Section 4.1, action-stitching methods at most learn the shortest path that \textbf{contained} in the datset. This is suboptimal especially when there does not exist one complete path starting from the start location to the goal location in the offline dataset.  

How can we do beyond action-stitching? One way is to use RL-based methods, by querying an accurate $Q^{\pi}$, we can get a different-yet-optimal action $a_{\pi}$, but $Q^{\pi}$ is hard to estimate.
Another way is like what \texttt{POR} did, we learn an out-of-distribution state indicator, i.e., the guide-policy $g$, to guide the policy to the optimal next state. If the execute-policy can generalize well, it will also output a different yet optimal action $a=\arg \max_{a} \pi(a | s, g(s))$.

\section{More Related Work} 
% Goal-conditioned RL, Hirrecical RL, Decoupled policy optimization in IL.
Our work decouples the state-to-action policy into two modules, i.e., the guide-policy and the execute-policy. The execute-policy is actually an inverse dynamics model, which has been widely used in various ways in sequential decision-making.
% the inverse dynamics model and the state transition planner. Both modules have been widely used by many pre- vious works on RL and IL tasks. 
In exploration, inverse dynamics can be used to learn representations of the controllable aspects of the state \citep{pathak2017curiosity}. In imitation learning, \citep{torabi2018behavioral} and \citep{liu2022plan} train an inverse dynamics model to label the state-only demonstrations with inferred actions. \citep{christiano2016transfer} use inverse dynamics models to translate actions taken in a simulated environment to the real world.

Recently, there has been an emergence of work \cite{srivastava2019training,ghosh2020learning} highlighting the connection of imitation learning and reinforcement learning. Specifically, rather than learn to map states and actions to reward, as is typical in reinforcement learning, \citep{srivastava2019training} trains a model to predict actions given a state and an outcome, which could be the amount of reward the agent is to collect within a certain amount of time. \citep{ghosh2020learning} uses a similar idea, predicting actions conditioned on an initial state, a goal state, and the amount of time left to achieve the goal. 
These methods are perhaps the closest work to our algorithm, however, we study the offline setting and motivate the usage of an inverse dynamics model from a different perspective (i.e., state-stitching).

% From the task perspective, they try to exactly match an expert demonstration sequence, with a binary classifier judging whether each goal is reached. 
% Kimura et al. (2018) utilized a state transition predictor to fit the state transi- tion probability in the expert data, which is further used to compute a predefined reward function. 
% Liu et al. (2020) con- structed a policy prior using the inverse dynamics and the state transition predictor, but the policy prior is trained in a supervised learning style and only used for regularizing the policy network. However, as shown in this paper, the policy can be exactly decoupled as these two parts without keeping an extra policy, where the state planner is optimized through policy gradient.

\section{Proof}
\label{sec: proof}
In this section, we provide the proof of Theorem \ref{theorem:ljx}.
% \subsection{Proof of Theorem \ref{theorem:ljx}}

\begin{proof}
We can rewrite the LHS of Eq.(\ref{equ:single_step_bound}) as
\begin{equation*}
\begin{aligned}
    \|\pi(s,g(s))-a^*\|
    &= \|\pi(s,g(s))-\pi(s,s')+\pi(s,s') - a + a -a_g+a_g-a^*\|\\
    &\le \|\pi(s,g(s))-\pi(s,s')\| + \|\pi(s,s')-a\| + \|a-a_g\| + \|a_g-a^*\|\ \ {(\rm Triangle)}\\
    &\le L_2\|g(s)-s'\|+\epsilon+L_1\|g(s)-s'\| +\|a_g-a^*\| \ {(\rm Assumption 1\&2)} \\
    &\le \underbrace{(L_1+L_2)\|g(s)-s'\|}_{l_1}+\underbrace{\|a_g-a^*\|}_{l_2}+\underbrace{\epsilon}_{l_3}
\end{aligned}
\end{equation*}
\end{proof}

\section{Experimental Details} \label{appendix: exp}
In this section, we provide the experimental details of our paper. 
% We use the following hardware and software for our training: 
    
% \begin{itemize}[nosep]
%     \item GPUs: NVIDIA GeForce RTX 3080Ti
%     \item Python 3.7
%     \item Pytorch 1.10.0 \cite{paszke2019pytorch}
%     \item Gym 0.23.1 \cite{brockman2016openai}
%     \item MuJoCo 2.1.4 \cite{mujoco}
%     \item mujoco-py 2.1.2.14
% \end{itemize}

% \subsection{Minigrid experiments}
% Regarding to Minigrid the toy case experiment. 

% \noindent \textbf{Implementation details} \quad

\subsection{D4RL Experiments} \label{appendix: exp d4rl}
\noindent \textbf{Data collection} \quad
The datasets in D4RL have been generated as follows: \texttt{random}: roll out a randomly initialized policy for 1M steps. \texttt{expert}: 1M samples from a policy trained to completion with SAC \cite{haarnoja2018soft}. \texttt{medium}: 1M samples from a policy trained to approximately 1/3 the performance of the expert. \texttt{medium-replay}: replay buffer of a policy trained up to the performance of the medium agent. \texttt{medium-expert}: 50-50 split of medium and expert data. For all datasets we use the v2 version.

\noindent \textbf{Implementation details} \quad 
Our implementation of 10\%BC is as follows, we first filter the top 10 $\%$ trajectories in terms of the trajectory return, and then run behaviour cloning on those filtered data. 
The hyperparameters of \texttt{POR} are present in Table \ref{table: hyperparams} and \ref{table: hyperparams more}.
We use target networks for the $V$-function and use clipped double $V$-learning (take the minimum of two $V$-functions) for all updates.
In some MuJoCo tasks, we add Layer Normalization \cite{ba2016layer} to $V$-networks to help stabilize training.
% We normalize state to $[-1, 1]$ \citep{stable-baselines3} to reduce the prediction error of the guide-policy, it can be deemed as an na\"ive method of representation learning. 

\begin{table}[htb]
\centering
\caption{General hyperparameters for \texttt{POR} and \texttt{POR-sparse}.}
\vspace{5pt}
\resizebox{0.7\linewidth}{!}{
\begin{tabular}{cll}
\toprule
& Hyperparameter & Value \\
\midrule
\multirow{9}{*}{Architecture}         
                                      & Value network hidden dim    & 256        \\
                                      & Value network hidden layers & 2          \\
                                      & Value network activation function & ReLU \\
                                      & Guide-policy hidden dim    & 256       \\
                                      & Guide-policy hidden layers & 2          \\
                                      & Guide-policy activation function & ReLU \\
                                      & Execute-policy hidden dim     & 1024        \\
                                      & Execute-policy hidden layers  & 2          \\
                                      & Execute-policy activation function & ReLU \\
                                      \midrule
\multirow{6}{*}{Hyperparameters} & Optimizer & Adam~\citep{adam} \\
                                      & Value netowrk learning rate & 1e-4 \\
                                      & Target Value netowrk moving average & 0.05 \\
                                      & Mini-batch size      & 256 \\
                                      & Discount factor      & 0.99 \\
                                      & Normalize  & False \\
                                    %   & Guide-policy learning objective  & (\ref{eq: update g two}) \\
                                    %   & Policy update frequency & 2 \\
                                    %   & Policy noise clip & (-5, 2) \\
\bottomrule
\label{table: hyperparams}
\end{tabular}
}
\end{table} %

\begin{table}[h]
\centering
\caption{Per-environment hyperparameters used for \texttt{POR}}
\resizebox{0.9\linewidth}{!}{
\begin{tabular}{l||rrrr} 
\toprule
Env                          & Guide-policy learning objective & $\tau$ & $\alpha$ & Policy learning rate \\ 
\midrule
antmaze-umaze-v2             & (\ref{eq: update g two}) & 0.9 & 10.0 & 1e-3 \\
antmaze-umaze-diverse-v2     & (\ref{eq: update g two}) & 0.9 & 10.0 & 1e-3 \\
antmaze-medium-play-v2       & (\ref{eq: update g two}) & 0.9 & 10.0 & 1e-4 \\
antmaze-medium-diverse-v2    & (\ref{eq: update g two}) & 0.9 & 10.0 & 1e-4 \\
antmaze-large-play-v2        & (\ref{eq: update g two}) & 0.9 & 10.0 & 1e-4 \\
antmaze-large-diverse-v2     & (\ref{eq: update g two}) & 0.9 & 10.0 & 1e-4 \\
\midrule
halfcheetah-medium-v2        & (\ref{eq: update g two})  & 0.5 & 3.0 & 1e-3 \\
hopper-medium-v2             & (\ref{eq: update g})  & 0.7 & 50.0 & 1e-3 \\
walker2d-medium-v2           & (\ref{eq: update g two})  & 0.5 & 3.0 & 1e-3 \\
halfcheetah-medium-replay-v2 & (\ref{eq: update g two})  & 0.5 & 3.0 & 1e-3 \\
hopper-medium-replay-v2      & (\ref{eq: update g})  & 0.7 & 50.0 & 1e-3 \\
walker2d-medium-replay-v2    & (\ref{eq: update g two})  & 0.5 & 3.0 & 1e-3 \\
halfcheetah-medium-expert-v2 & (\ref{eq: update g two})  & 0.5 & 3.0 & 1e-3 \\
hopper-medium-expert-v2      & (\ref{eq: update g})  & 0.7 & 5.0 & 1e-3 \\
walker2d-medium-expert-v2    & (\ref{eq: update g two})  & 0.5 & 3.0 & 1e-3 \\
\bottomrule
\label{table: hyperparams more}
\end{tabular}
}
\end{table}

\begin{table}[h]
\centering
\caption{Per-environment hyperparameters used for \texttt{POR-sparse}}
\resizebox{0.8\linewidth}{!}{
\begin{tabular}{l||rrrr} 
\toprule
Env                          & Guide-policy learning objective & $\tau$ & $\alpha$ & Policy learning rate \\ 
\midrule
antmaze-umaze-v2             & (\ref{eq: update g two}) & 1.0 & 10.0 & 1e-3 \\
antmaze-umaze-diverse-v2     & (\ref{eq: update g two}) & 0.5 & 10.0 & 1e-3 \\
antmaze-medium-play-v2       & (\ref{eq: update g two}) & 0.1 & 10.0 & 1e-4 \\
antmaze-medium-diverse-v2    & (\ref{eq: update g two}) & 0.1 & 10.0 & 1e-4 \\
antmaze-large-play-v2        & (\ref{eq: update g two}) & 0.1 & 10.0 & 1e-4 \\
antmaze-large-diverse-v2     & (\ref{eq: update g two}) & 0.3 & 10.0 & 1e-4 \\
\bottomrule
\label{table: hyperparams more sql}
\end{tabular}
}
\end{table}

\subsection{Additional Suboptimal Data Experiments} \label{appendix: exp add data} 
\noindent \textbf{Data collection and settings} \quad 

In this experiment, we use different part of \texttt{medium-replay} datasets as $\mathcal{D}_{e}$ and $\mathcal{D}_{o}$.
More specific, we use $30\%$ to $70\%$ transitions to constitute $\mathcal{D}_{e}$ and use $20\%$ to $80\%$ transitions to constitute $\mathcal{D}_{e} \cup \mathcal{D}_{o}$.

\noindent \textbf{Implementation details} \quad We use the same hyperparamters of \texttt{POR}, shown in Table \ref{table: hyperparams}. Our implementations of TD3+BC\footnote{\url{https://github.com/sfujim/TD3_BC}}~\citep{fujimoto2018addressing}, CQL\footnote{\url{https://github.com/aviralkumar2907/CQL}}~\citep{kumar2020conservative} is from the author-provided implenmentation from Github, and we keep all parameters the same to the author-provided implementation.

\begin{table}[htb]
\caption{The hyperparameters of CQL in additional-data experiments.}
\centering
\vspace{5pt}
\resizebox{0.7\linewidth}{!}{
\begin{tabular}{cll}
\toprule
& Hyperparameter & Value \\
\midrule
\multirow{6}{*}{Architecture}         & Critic hidden dim    & 256        \\
                                      & Critic hidden layers & 3          \\
                                      & Critic activation function & ReLU \\
                                      & Actor hidden dim     & 256        \\
                                      & Actor hidden layers  & 3          \\
                                      & Actor activation function & ReLU \\
\midrule
\multirow{12}{*}{CQL Hyperparameters} & Optimizer & Adam~\citep{adam}    \\
                                      & Critic learning rate & 3e-4       \\
                                      & Actor learning rate  & 1e-4      \\
                                      & Mini-batch size      & 256        \\
                                      & Discount factor      & 0.99       \\
                                      & Target update rate   & 5e-3   \\
                                      & Target entropy       & -1 $\cdot$ Action Dim \\
                                      & Entropy in Q target  & True  \\ 
                                      & Lagrange             & False      \\
                                    %   & Pre-training steps   & 40e3       \\
                                      & Num sampled actions (during eval) & 10 \\
                                      & Num sampled actions (logsumexp) & 10 \\
                                      & $\alpha$                & 10         \\
\bottomrule
\end{tabular}
}
\label{table:cql_hyp}
\end{table}

\begin{table}[htb]
\centering
\caption{The hyperparameters of TD3+BC in additional-data experiments.}
\vspace{5pt}
\resizebox{0.6\linewidth}{!}{
\begin{tabular}{cll}
\toprule
& Hyperparameter & Value \\
\midrule
\multirow{6}{*}{Architecture}         & Critic hidden dim    & 256        \\
                                      & Critic hidden layers & 2          \\
                                      & Critic activation function & ReLU \\
                                      & Actor hidden dim     & 256        \\
                                      & Actor hidden layers  & 2          \\
                                      & Actor activation function & ReLU \\
                                      \midrule
\multirow{10}{*}{TD3+BC Hyperparameters} & Optimizer & Adam~\citep{adam} \\
                                      & Critic learning rate & 3e-4 \\
                                      & Actor learning rate  & 3e-4 \\
                                      & Mini-batch size      & 256 \\
                                      & Discount factor      & 0.99 \\
                                      & Target update rate   & 5e-3 \\
                                      & Policy noise         & 0.2 \\
                                      & Policy noise clipping & (-0.5, 0.5) \\
                                      & Policy update frequency & 2 \\
                                      & $\alpha$             & 2.5 \\
\bottomrule
\end{tabular}
}
\label{table:td3_hyp}
\end{table} %

\subsection{Four-room Experiments} \label{appendix: exp fou room}
\noindent \textbf{Environment settings} \quad 
We use the continuous variant of the classic four-room environment from \citep{janner2021offline}. 
This continuous variant of four-rooms is basically the same as the traditional classic four-room environment in the environment design. There are $19 \times 19$ grids that consist of four rooms with only one block channel with neighboring rooms. The goal of this environment is to make the agent travel from one location to another different location. Those tasks are challenging as their reward is extremely spare, they need the agent to have the ability to explore efficiently through the whole state space. The action and observation space are shown in Table \ref{table: overall action observation space}. The first and second dimension of the action space represents the distance to travel on the $x$-axis and $y$-axis, respectively. The first and second dimension of the observation space represents the coordinate on the $x$-axis and $y$-axis, respectively.  

\begin{table}[htb]
\centering
\caption{The action and observation space in the continuous \texttt{Four-room} environment.}
\vspace{5pt}
\resizebox{0.5\linewidth}{!}{
\begin{tabular}{cc} 
\toprule
Action space      & Box(-0.1, 0.1, (2,), float32)  \\
Observation space & Box(-18, 18, (2,), float32)    \\
\midrule
Action dimension      & 2  \\
Observation dimension & 2    \\
\bottomrule
\end{tabular}
}
\label{table: overall action observation space}
\end{table}

In task A (\texttt{Four-room}), only reaching the goal will give the agent a reward of $1$, otherwise the agent will get $0$ reward. 
When the agent reaches the target or the agent takes over 500 steps, the environment will terminate. 
In task B (\texttt{Four-room-river}), the agent will receive $-1$ reward if it falls into the river, and the environment will be terminated. Also, the agent will get $1$ reward only when it reaches the goal.
In task C (\texttt{Four-room-key}), the agent gets $1$ reward only when it reaches the goal \textbf{and} gets the key. Falling into the river will also give the agent $-1$ reward.

\noindent \textbf{Data collection and implementation details} \quad 
The training data consists of
trajectories collected by a goal-reaching controller with the start and end locations sampled randomly at non-wall locations. To make the data more diverse, we also add trajectories from a random policy. We collect 100,000 transitions for each task. Figure \ref{table: four-room hyperparamters} shows the training hyperparameters of \texttt{POR} in the three four-room environments. 

\begin{table}[htbp]
\centering
\caption{The hyperparameters of \texttt{POR} in continuous \texttt{Four-room} environments. }
\vspace{5pt}
\resizebox{0.7\linewidth}{!}{
\begin{tabular}{cll}
\toprule
& Hyperparameter & Value \\
\midrule
\multirow{9}{*}{Architecture}         
                                      & Value network hidden dim    & 64        \\
                                      & Value network hidden layers & 2          \\
                                      & Value network activation function & ReLU \\
                                      & Guide-policy hidden dim    & 64       \\
                                      & Guide-policy hidden layers & 2          \\
                                      & Guide-policy activation function & ReLU \\
                                      & Execute-policy hidden dim     & 64        \\
                                      & Execute-policy hidden layers  & 2          \\
                                      & Execute-policy activation function & ReLU \\
                                            \midrule
\multirow{9}{*}{Training Hyperparameters} & Optimizer & Adam~\citep{adam} \\
                                      & Value network learning rate & 1e-4 \\
                                      & Target V moving average & 0.05 \\
                                      & Guide-policy learning rate  & 1e-4 \\
                                      & Execute-policy learning rate  & 1e-4 \\
                                      & Mini-batch size      & 256 \\
                                      & Discount factor      & 0.99 \\
                                    %   & Policy update frequency & 2 \\
                                    %   & Policy noise clip & (-5, 2) \\
                                      & Normalize  & False \\
                                      & Guide-policy learning objective  & (\ref{eq: update g two}) \\
                                      & $\tau$       &  0.9 \\
                                      & $\alpha$       &  10.0 \\
\bottomrule
\label{table: four-room hyperparamters}
\end{tabular}
}
\end{table} %

% \section{More Experiments}
% \lipsum[1-3]
% \clearpage
\section{Learning Curves} \label{appendix: learning curve}
In this section, we provide the learning curves of our experiments in the main paper.

\begin{figure}[htbp]
\centering
\includegraphics[width=1.0\columnwidth]{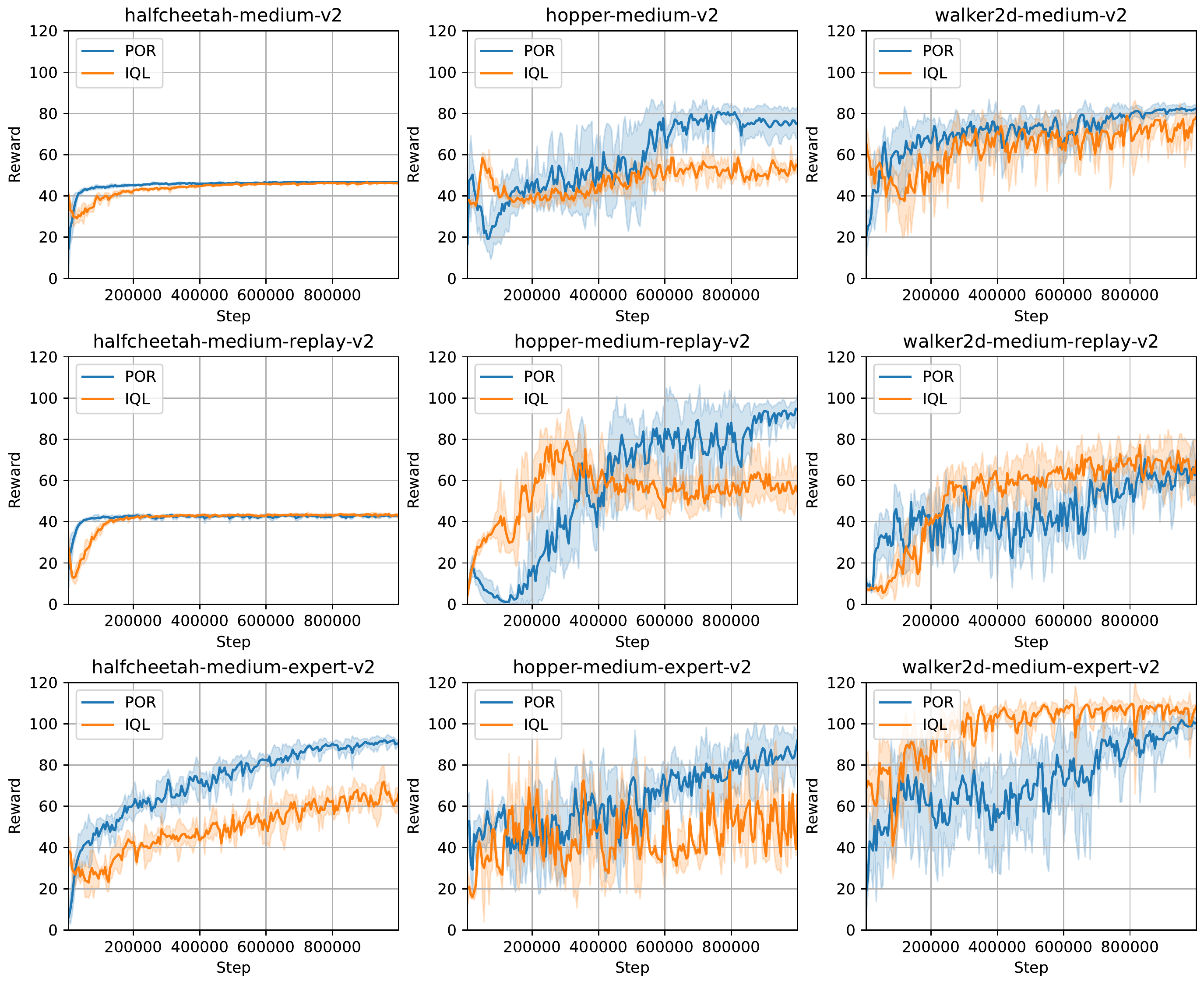} 
\caption{Learning curves of \texttt{POR} against our implemented \texttt{IQL} (using the same codebase and network structures) on D4RL MuJoCo datasets.}
\label{fig: learning curve locomotion}
\end{figure}

\begin{figure}[htbp]
\centering
\includegraphics[width=1.0\columnwidth]{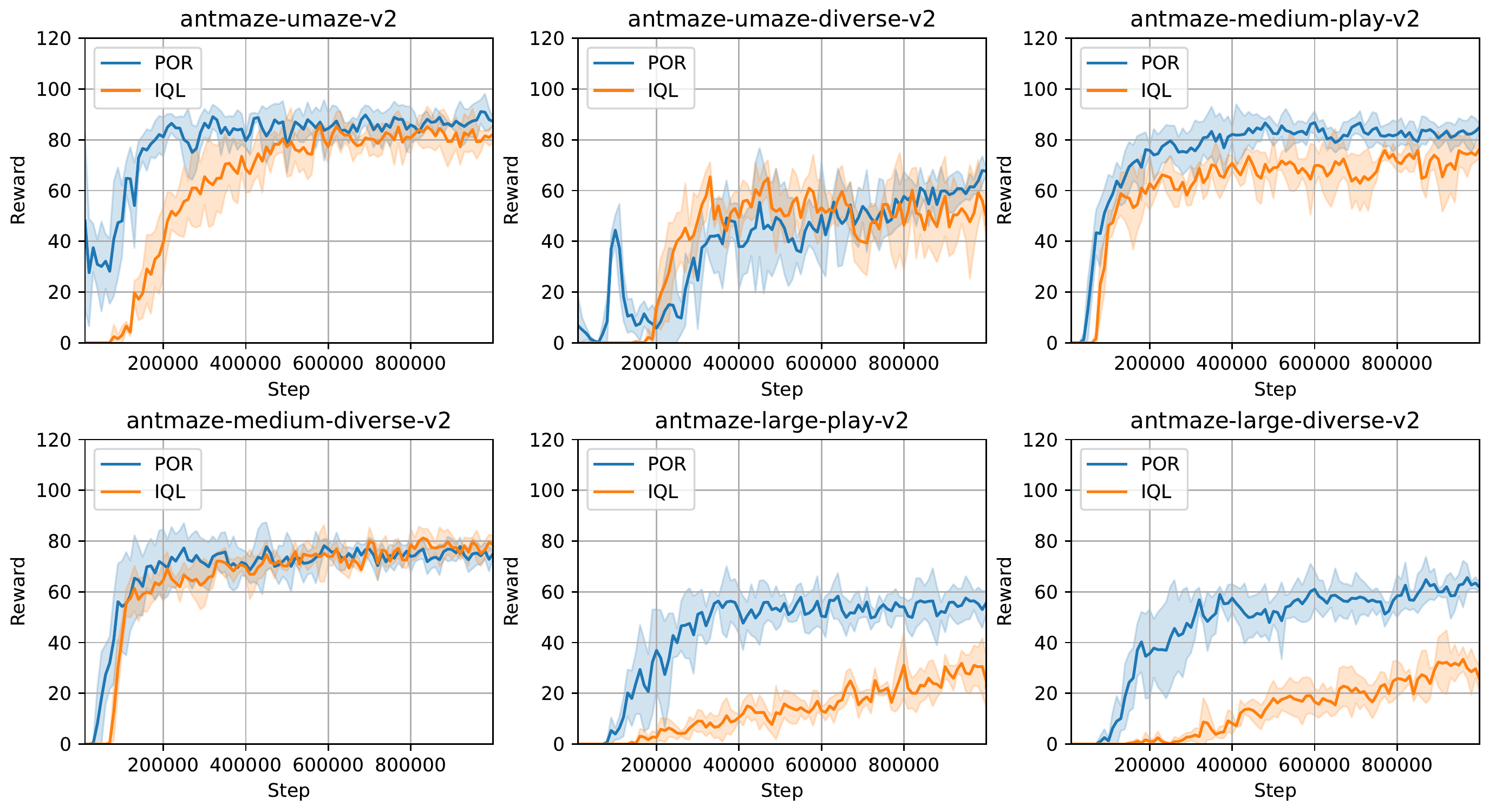} 
\caption{Learning curves of \texttt{POR} against our implemented \texttt{IQL} (using the same codebase and network structures) on D4RL Antmaze datasets.}
\label{fig: learning curve antmaze}
\end{figure}

\begin{figure}[htbp]
\centering
\includegraphics[width=1.0\columnwidth]{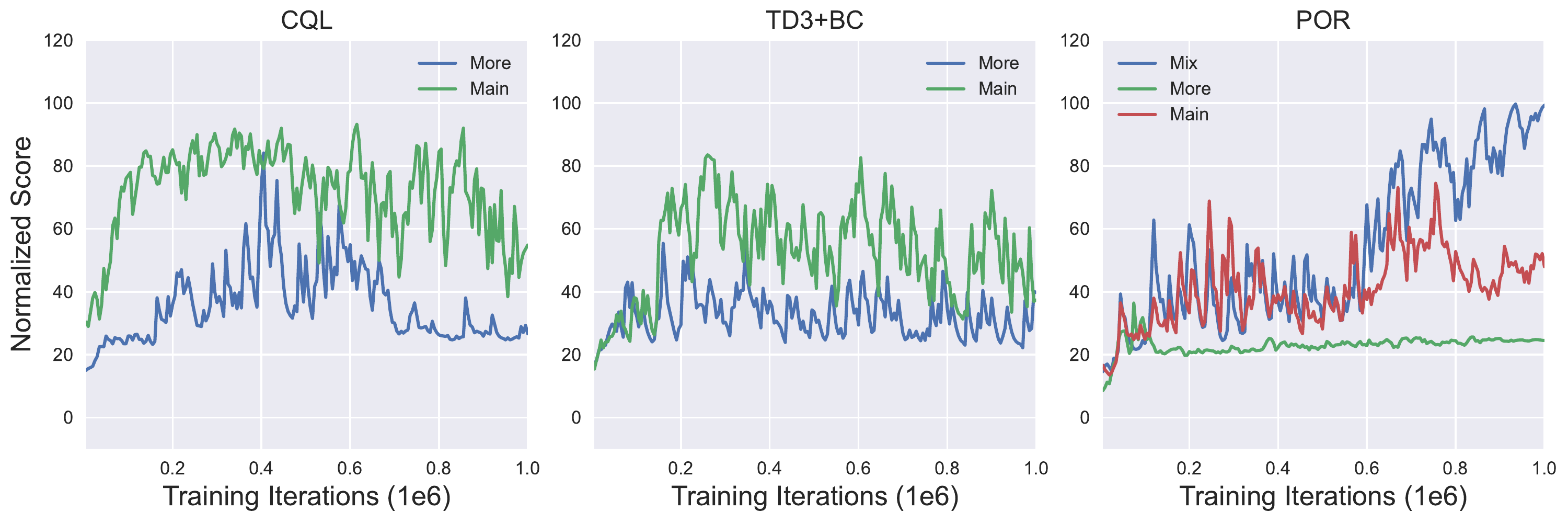} 
\caption{Learning curves of \texttt{POR} in additional-data experiments on hopper-medium-replay-v2 datasets.}
\label{fig_add_data_hopper}
\end{figure}

\begin{figure}[htbp]
\centering
\includegraphics[width=1.0\columnwidth]{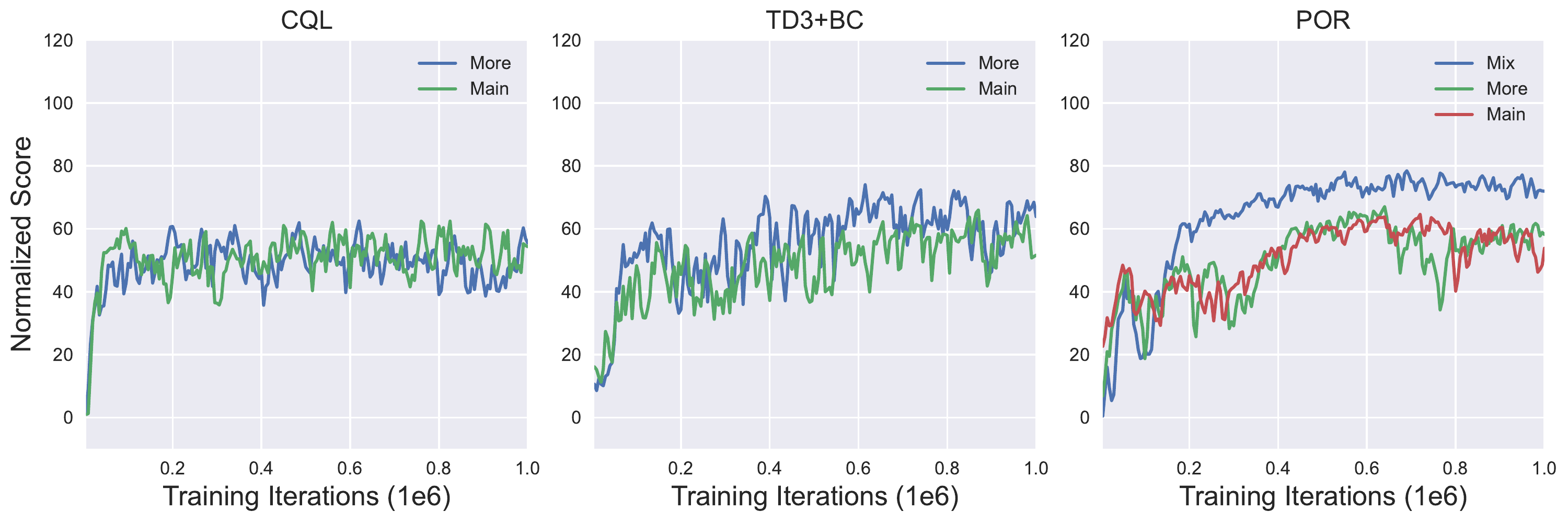} 
\caption{Learning curves of \texttt{POR} in additional-data experiments on walker2d-medium-replay-v2 datasets.}
\label{fig_add_data_walker2d}
\end{figure}

\section{Ablation Study on the Execute-Policy}
\label{appendix: ablation study}

We also present an ablation study on the execute-policy, with the aim to answer the following two questions: 
1) How does network capacity affect the performance of the execute-policy? 
2) How does the performance change if we use partial "good" data, instead of full data, to train the execute policy?
% to study the relationship between the hyperparameters and our pro- posed method.

To answer the first question, we compare two choices of network size: a big network with $(512, 512)$ hidden units and a small network with $(128, 128)$ hidden units.
To answer the second question, we compare with the choice that only selects a subset of the dataset to train the execute policy. Concretely, we choose the top $X\%$ trajectories in the dataset, ordered by episode returns. We sweep $X$ over $[10, 25, 40]$ and choose the best score.
We give the mean scores of AntMaze (\texttt{A}) and MuJoCo (\texttt{M}) datasets in Table \ref{table: generlization}. It can be seen that adopting a big network consistently gives a better performance, which is also found in \cite{emmons2021rvs}. Using partial data will result in a less-performed policy, especially on AntMaze datasets. Note that in MuJoCo datasets, the performance didn't drop too much. This is because MuJoCo datasets don't require the compositionality ability, using partial trajectories could already achieve high scores.

\begin{table}[htb]
\caption{Ablation study of the execute-policy on network capacity and dataset size.}
\centering
\vspace{5pt}
\begin{tabular}{ccc} 
\toprule
             & Big Network & Small Network  \\ 
\midrule
Full $\mathcal{D}$   & \texttt{A}: 76.2, \texttt{M}: 65.7  & \texttt{A}: 63.8, \texttt{M}: 57.0  \\
Partial $\mathcal{D}$  & \texttt{A}: 52.4, \texttt{M}: 60.9  & \texttt{A}: 41.5, \texttt{M}: 56.7  \\
\bottomrule
\end{tabular}
\label{table: generlization}
\end{table}

\end{document}